\documentclass[pmlr]{jmlr}

\RequirePackage{graphicx}
 \usepackage{booktabs}
\usepackage{longtable}
\usepackage{float} 
\makeatletter
\def\set@curr@file#1{\def\@curr@file{#1}} 
\makeatother
\usepackage[load-configurations=version-1]{siunitx} 

\usepackage{multirow}
\theorembodyfont{\upshape}
\theoremheaderfont{\scshape}
\theorempostheader{:}
\theoremsep{\newline}

\jmlrvolume{298}
\jmlryear{2025}
\jmlrworkshop{Machine Learning for Healthcare}

\title[Uncertainty-Aware Prediction of Parkinson's Disease Medication Needs]{Uncertainty-Aware Prediction of Parkinson's Disease Medication Needs: A Two-Stage Conformal Prediction Approach}

\author
       {\Name{Ricardo Diaz-Rincon}
       \Email{rdiazrincon@ufl.edu}\\ 
       \addr Department of Neuroscience\\
       \addr University of Florida\\
       \vspace{-5pt}
       \AND
       \Name{Muxuan Liang}
       \Email{muxuan.liang@ufl.edu}\\ 
       \addr Department of Biostatistics\\
       \addr University of Florida\\
       \vspace{-5pt}
        \AND
       \Name{Adolfo Ramirez-Zamora}
       \Email{adolfo.ramirez-zamora@neurology.ufl.edu}\\ 
       \addr Department of Neurology\\
       \addr University of Florida\\
       \vspace{-5pt}
        \AND
       \Name{Benjamin Shickel}
       \Email{shickelb@ufl.edu}\\ 
       \addr Department of Medicine\\
       \addr University of Florida\\}

\begin{document}

\maketitle
\vspace{-40pt}
\begin{abstract}
  Parkinson's Disease (PD) medication management presents unique challenges due to heterogeneous disease progression, symptoms,  and treatment response. Neurologists must balance symptom control with optimal dopaminergic medication dosing based on functional disability while minimizing risks of side effects. This balance is crucial as inadequate or abrupt changes can cause levodopa-induced dyskinesia (LID), wearing off, and neuropsychiatric side effects, significantly reducing quality of life. Current approaches rely on trial-and-error decision-making without systematic predictive methods. Despite machine learning advances in medication forecasting, clinical adoption remains limited due to reliance on point predictions that do not account for prediction uncertainty, undermining clinical trust and utility. To facilitate trust, clinicians require not only predictions of future medication needs but also reliable confidence measures. Without quantified uncertainty, medication adjustments risk premature escalation to maximum doses or prolonged periods of inadequate symptom control. To address this challenge, we developed a conformal prediction framework anticipating medication needs up to two years in advance with reliable prediction intervals and statistical guarantees. Our approach addresses zero-inflation in PD inpatient data, where patients maintain stable medication regimens between visits. Using electronic health records data from 631 inpatient admissions at University of Florida Health (2011-2021), our novel two-stage approach identifies patients likely to need medication changes, then predicts required levodopa equivalent daily dose adjustments. Our framework achieved marginal coverage while significantly reducing prediction interval lengths compared to traditional approaches, providing precise predictions for short-term planning and appropriately wider ranges for long-term forecasting, matching the increasing uncertainty in extended projections. By quantifying uncertainty in medication needs, our approach enables evidence-based decisions about levodopa dosing and medication adjustments, potentially optimizing symptom control while minimizing side effects and improving patients' quality of life. 
\end{abstract}

\section{Introduction}

Parkinson's Disease (PD) presents significant medication management challenges due to its heterogeneous nature, with patient-specific variability in disease progression and treatment response \citep{Fox2018}. Neurologists face the delicate task of balancing symptom control against potential side effects when adjusting dopamine replacement therapies (DRT), particularly levodopa, the gold-standard medication for treating PD \citep{Armstrong2020}. Inadequate dosing or abrupt changes in medication regimens can lead to profound complications such as levodopa-induced dyskinesia (LID), cognitive or psychiatric side effects, while insufficient dosing may result in poor symptom control and reduced quality of life. This balance is traditionally achieved through subjective assessments and trial-and-error approaches, highlighting the need for systematic, data-driven decision-support tools. 

The levodopa equivalent daily dose (LEDD) provides a standardized metric for comparing various antiparkinsonian medications by converting different drugs and formulations to a common scale \citep{Tomlinson2010}. Monitoring and adjusting LEDD over time is crucial for optimizing clinical outcomes, yet current approaches offer limited assistance in forecasting medication adjustments over time. This limitation is particularly critical given that treatment responses vary significantly between patients and over time, making point predictions alone insufficient for clinical decision-making due to their lack of uncertainty quantification, potentially leading to overconfident treatment decisions that fail to account for the inherent variability in patient responses.

Electronic health records (EHR) data contain valuable but underutilized information about medication use, clinical outcomes, and disease progression patterns. Although machine learning approaches have shown promise in leveraging these data for predictive modeling, their potential clinical utility is limited when predictions directly inform high-stakes treatment decisions without quantifying uncertainty. Even when prediction models achieve high accuracy, their point estimates do not capture the inherent variability in patient responses, potentially leading to overconfident and potentially harmful AI-augmented treatment decisions. For trusted clinical adoption, prediction systems must not only be accurate, but also provide reliable measures of uncertainty that clinicians can incorporate into their decision-making processes and mental models of disease \citep {Begoli2019}.

Conformal prediction (CP) is a powerful statistical framework that addresses these limitations by providing distribution-free prediction intervals with guaranteed coverage under minimal assumptions \citep{Angelopoulos2020, Vovk2005}. Unlike traditional uncertainty estimation methods that rely on distributional assumptions, conformal prediction can adapt to the complex, non-standard patterns of PD progression, while maintaining rigorous statistical guarantees about prediction reliability.  

For instance, rather than simply predicting that a patient will need a 15\% increase in medication dosage at their next hospital visit, conformal prediction provides a range (e.g., 10-20\% increase) with a statistical guarantee that the true required medication adjustment falls within this interval with a pre-specified confidence level (e.g., 90\%). This approach is particularly valuable in PD management, where medication decisions carry significant consequences for symptom control and quality of life, making uncertainty quantification essential to guide medication management and clinical decisions. 

Our research introduces a novel application of conformal prediction to PD medication management, providing reliable intervals for forecasted LEDD changes. Our approach uniquely addresses the zero-inflation challenge in PD inpatient medication data (i.e., many patients maintain stable medication regimens between visits leading to no change in LEDD, represented as a target value of zero) by developing a two-stage modeling approach that first identifies patients likely to need medication adjustments, then predicting the magnitude of change in medication. This allows neurologists and movement disorder specialists to directly visualize the degree of uncertainty in each prediction, i.e., narrower intervals suggest higher confidence in specific dosage adjustments, while wider intervals signal greater variability and increased uncertainty in potential outcomes, prompting more cautious or conservative approaches to medication changes. By combining machine learning with rigorous uncertainty quantification, we aim to bridge the gap between clinical expertise and data-driven decision-making, ultimately advancing personalized medication management and improving patient outcomes in PD care.



\subsection*{Generalizable Insights about Machine Learning in the Context of Healthcare}
Our work provides several insights applicable to machine learning in healthcare beyond the management of Parkinson's Disease medications. First, we demonstrate how to effectively handle zero-inflation in longitudinal healthcare data, a common challenge in medical applications through a two-stage modeling approach.
Additionally, we highlight how uncertainty quantification methods can be integrated into clinical decision support systems to provide statistically valid prediction intervals rather than point estimates, conveying meaningful information about patient trajectories, serving as an interpretable measure guiding treatment strategies. 
Furthermore, our comparative analysis reveals the critical importance of calibration quality across different time horizons, demonstrating that prediction methods should be systematically evaluated for their performance across multiple forecasting periods rather than assuming uniform reliability.
Finally, we demonstrate the inherent methodological tension between providing more precise (narrower) intervals and maintaining statistical reliability, a critical trade-off in medical domains where patient safety relies on trustworthy uncertainty estimates.

\section{Background and Related Work}

\subsection{Parkinson’s Disease Medication Management}  
Parkinson's Disease presents significant challenges for medication management due to its heterogeneous nature and variable progression \citep{Jankovic1990, Fereshtehnejad2015}. Levodopa remains the gold standard medication \citep{Cotzias1967} but it is associated with long-term complications related to disease progression, severity, and levodopa dosing, namely end of dose wearing off and levodopa-induced dyskinesias (LID), which are debilitating involuntary movements that significantly impact quality of life \citep{Olanow2006, Jenner2008}. To standardize comparison across diverse medication regimens, the LEDD has been established as a metric that converts different antiparkinsonian medications to their levodopa-equivalent values \citep{Tomlinson2010, Joshi2010, Kukkle2024}. Current medication management relies on periodic clinical assessments (typically every 3-6 months), subjective patient reporting, and specialist expertise \citep{Postuma2015, Goetz2003, Goetz2008}. This approach often fails to capture the full complexity of disease progression and treatment response, resulting in suboptimal medication management \citep{Dorsey2020}, particularly in inpatient settings where medication errors occur in over 70\% of PD admissions \citep{Aminoff2011, Gerlach2011}. These challenges underscore the need for improved methods to predict medication requirements with appropriate uncertainty quantification. The pervasiveness and serious consequences of these medication management challenges underscore the need for data-driven approaches that can provide reliable predictions of medication needs with appropriate uncertainty quantification. An extended PD background can be found in Appendix~\ref{sec:extended_pd}.

\subsection{Machine Learning in PD}  

Previous computational approaches to PD management have primarily focused on diagnosis, subtype classification, and motor symptom assessment rather than medication prediction. \citep{Belyaev2023} utilized machine learning and electroencephalography (EEG) for diagnosis and monitoring of PD. Studies by \citep{Lusk2023} and \citep{Palacios2011} focused on using EHR data for statistical analysis and risk factor identification.  Recent work by \citep{Yuan2021} focused on the prediction of PD based on prodromal symptoms. Similarly, \citep{Latourelle2017} employed machine learning to predict disease progression using genetic data, though focused on motor symptoms rather than medication needs. These studies highlight a significant gap in applying machine learning to PD research, especially in medication adjustment. 

Prior applications of machine learning to support PD medication management have focused on individual medications or binary response classifications rather than utilizing the clinically validated LEDD metric that allows standardized comparison across different medication regimens. Even among the studies that incorporated LEDD, most were limited to classification, clustering, or narrow contexts like post-surgical adjustments rather than providing comprehensive longitudinal forecasting of medication needs throughout disease progression \citep{Watts2021}. Most critically, none of these studies incorporated uncertainty quantification, a critical element for trusted clinical decision support in a heterogeneous condition like PD. 

\citep{Riasi2024} utilized a type of recurrent neural network approach, the Long Short-Term Memory (LSTM) model to predict dosages of five medication types for PD (levodopa, dopamine agonists, MAO-B inhibitors, COMT inhibitors, and amantadine) using the Parkinson’s Progression Markers Initiative (PPMI) dataset \citep{Marek2011}. The authors incorporated patient histories for temporal modeling and most medications used in the treatment of PD. However, they focused on individual drug dosages rather than using the LEDD, which provides a standardized comparison across different PD medications. Additionally, the authors provided point estimates without uncertainty measures. Similarly, \citep{Gutowski2023} utilized neural networks to predict treatment responses and optimize carbidopa-levodopa dosing schedules, but relied on simulated data, potentially limiting clinical utility and failing to accurately represent real-world disease progression and medication use. Their work also focused exclusively on optimizing the timing and size of carbidopa-levodopa rather than taking advantage of the standardization of antiparkinsonian drugs that the LEDD provides. Likewise, no uncertainty quantification was performed in this study. 

Among studies incorporating LEDD, \citep{Salmanpour2023} used random forest models to predict levodopa changes across 5 years of PD treatment using the PPMI dataset. Although they calculated the LEDD, their predictive models primarily focused on raw levodopa dosages and fragmented the longitudinal trajectory by analyzing changes across specific time intervals (years 1-2, 2-3, etc.) rather than developing a unified forecasting model for future dose trajectories with quantified uncertainty. \citep{Shamir2015} utilized LEDD in their work on deep brain stimulation outcomes using support vector machines, naïve bayes, and random forest. While their inclusion of LEDD represents a step toward standardized medication quantification, their analysis was limited to post-surgical adjustments rather than comprehensive longitudinal medication outcomes. Their approach also lacked uncertainty estimates. \citep{Watts2021} explicitly used LEDD for clustering patients based on their medication regimens and predicting cluster membership using sensor data but were limited to classification into predetermined medication groups rather than forecasting future needs. 

Other approaches like \citep{Kim2021} utilized reinforcement learning to optimize medication combinations for symptom control using the PPMI dataset. While their model successfully identified regimens that minimized motor symptoms, it focused on immediate decision-making rather than long-term forecasting, analyzed individual drugs rather than comprehensively using the LEDD, and failed to provide uncertainty measures. \citep{Jatain2021} classified PD patients as "good" or "bad" levodopa responders using the Michael J. Fox Foundation Levodopa Response Trial Wearable dataset but offered only binary classification without addressing longitudinal progression or medication dosage prediction. 

Appendix~\ref{sec:lit_summary} shows a summary of the current state of the literature with Table~\ref{tab:lit_table} showing the differences between approaches. As signified by \citep{Begoli2019} in their review of clinical decision support systems, providing point estimates only, in heterogeneous conditions like PD, can lead to overconfident treatment decisions and potentially suboptimal outcomes. This gap underscores the need for methods that not only predict medication changes but also quantify prediction uncertainty in a clinically meaningful manner.

\subsection{Conformal Prediction} 

Conformal prediction, introduced by \citep{Vovk2005} provides a distribution-free framework for constructing prediction intervals with guaranteed coverage properties. Unlike traditional parametric approaches that require assumptions about data distributions, conformal prediction produces valid prediction regions regardless of the underlying distribution, making it particularly suitable for the non-standard patterns observed in PD progression \citep{Angelopoulos2023, Shafer2008, Angelopoulos2021}.

The core principle of conformal prediction involves using past prediction errors to calibrate future prediction intervals. For each new instance, the method computes a nonconformity score, measuring how ``unusual'' the instance is compared to previous observations. By selecting a desired coverage (e.g., 90\%), CP guarantees that the true value will fall within the predicted interval with that probability \citep{Angelopoulos2024}. This methodological advantage is especially valuable in PD medication management, where both overestimation and underestimation of treatment needs can significantly impact patient outcomes.  

Several methods of conformal prediction have been developed to address specific challenges in machine learning applications. The Naïve approach, as defined in \citep{Barber2021} constructs prediction intervals using in-sample residuals from the same data used to fit the model. Split conformal prediction divides the data into proper training and calibration sets, enhancing computational efficiency while maintaining coverage guarantees \citep{Papadopoulos2002}. 

More advanced techniques offer particular advantages for clinical applications like PD medication prediction. Cross validation (CV+), intricately related to cross-conformal prediction \citep{Vovk2015, Vovk2018}, leverages multiple training and calibration splits to reduce prediction interval lengths while preserving coverage guarantees, a great advantage when recommending precise medication adjustments \citep{Barber2021}. Jackknife+ employs a leave-one-out strategy to efficiently construct prediction intervals \citep{Barber2021} while jackknife+ after bootstrap (J+aB) builds upon this by combining bootstrap sampling with jackknife's efficiency \citep{Kim2020}. This approach produces intervals that are particularly robust when working with limited or heterogeneous data, making it well-suited for clinical datasets where patient variability is high, and sample sizes may be constrained \citep{Alaa2020}. 

These methodological innovations have enabled diverse healthcare applications. In clinical settings, \citep{Angelopoulos2020} applied conformal prediction to critical care outcomes, showing how prediction intervals with valid coverage can improve clinical decision-making. \citep{Shafer2008} explored its theoretical advantages for clinical decision support, highlighting how distribution-free guarantees can address the variability inherent in biological systems. Particularly relevant is work by \citep{Romano2019} on adaptive conformal inference, which addresses challenges in settings with distribution shifts. Additional healthcare applications include breast cancer survivability and early diagnosis \citep{Alnemer2016, Devetyarov2012}, stroke risk estimation \citep{Papadopoulos2017} and acute abdominal pain \citep{Papadopoulos2009, Papadopoulos2009b}. 

Despite these diverse applications and growing adoption in healthcare, there remains a lack of conformal approaches specifically tailored to zero-inflated data, an important characteristic of many clinical tasks, including PD medication management. By providing intervals with statistical guarantees rather than point estimates, conformal prediction directly addresses the practical needs of clinicians making high-stakes decisions amid significant uncertainty in medication needs and patient heterogeneity. In the following methodology section, we describe how to extend and apply this framework to address the specific challenges of PD medication management. 

\section{Methodology}

What distinguishes our approach is the novel application of conformal prediction to medication management in PD, combining structured EHR data with unstructured clinical narratives to provide reliable prediction intervals for LEDD changes. Our methodology addresses a key challenge in PD inpatient medication forecasting: zero-inflation, where a substantial proportion of patients maintain stable medication regimens between hospital visits despite disease progression, leading to no change in medications and, therefore, a zero-valued outcome. To overcome this limitation, we developed a two-stage modeling strategy that first identifies patients likely to require medication adjustments using a classification model, then predicts the magnitude of LEDD change for those patients only. This targeted approach not only improves prediction accuracy but also ensures that uncertainty quantification focuses on clinically actionable cases where medication changes are needed. 


\subsection{Data Source}

We utilized deidentified EHR data in the OMOP (Observational Medical Outcomes Partnership) common data model (CDM) format from inpatient admissions at University of Florida Health between 2011-2021. This dataset includes both structured clinical data and unstructured clinical narratives, providing a comprehensive view of patient care over multiple hospitalizations. 

\subsection{Patient Cohort} 
We performed initial data extraction using ICD-9 codes (332.0 for primary PD, 332.1 for secondary parkinsonism, 781.0, 781.2, 781.3 for related motor symptoms) and ICD-10 codes (G20, G20.X for primary PD; G21.X for secondary parkinsonism; G25.X for parkinsonian tremors; R25.X, R26.X, R27.0 for related motor symptoms) to identify potential PD cases. Then, we validated diagnoses through medication history (presence of antiparkinsonian medications) and clinical note review to confirm accurate PD cohort selection.  Our cohort comprises 1,456 initial patients identified through ICD codes, from which we validated 631 patients through rigorous review of PD diagnosis and inpatient medication administration records. Table~\ref{tab:tab_demographics} presents the demographic information of the study cohort. Appendix~\ref{sec:pd_meds}, Table~\ref{tab:pd_meds} shows administered PD medications by pharmacologic class.

\begin{table}[t]
\small
  \centering 
  \caption{Demographic Characteristics of the Study Cohort}
  \begin{tabular}{llll}
    \toprule
    
    \textbf{Characteristic} & \textbf{Category} & \textbf{Number} & 
    \textbf{Percentage (\%)} \\
    \multicolumn{2}{l}{Total Patients} & 631 & 100.00 \\
    \midrule
    \multicolumn{2}{l}{Mean Age, Years (SD)} & \multicolumn{2}{l}{77.98 (10.59)} \\
    \midrule
    \multirow{2}{*}{Gender} 
      & Male & 407 & 64.50 \\ 
      & Female & 224 & 35.50 \\ 
    \midrule
    \multirow{5}{*}{Race} 
      & White & 526 & 83.62 \\ 
      & Black & 63 & 10.01 \\ 
      & Other & 34 & 5.41 \\ 
      & Asian & 5 & 0.79 \\ 
      & Multiracial & 1 & 0.16 \\ 
    \midrule
    \multirow{2}{*}{Ethnicity} 
      & Not Hispanic & 598 & 95.07 \\ 
      & Hispanic & 31 & 4.93 \\ 
    \bottomrule
  \end{tabular}
  \label{tab:tab_demographics}
\end{table}

\subsection{Feature Engineering and LEDD Computation} 

\begin{figure}[H]
   \centering 
   \includegraphics[width=2.8in]{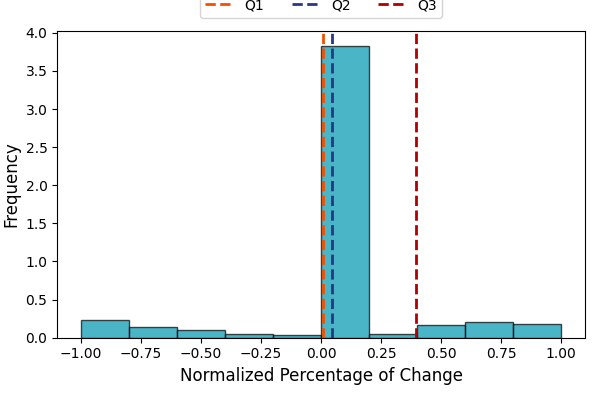}
   \caption{Distribution of normalized LEDD percentage change across patient visits. The peak centered at zero indicates approximately 75\% of observations showing no change in medication dosage between visits.} 
   \label{fig:normalized_pct_change} 
 \end{figure}

We extracted structured data including demographics (age, sex, race and ethnicity), medication records (drug name, dosage, frequency, administration route), hospital visit information (admission dates, length of stay, department), and clinical variables relevant to PD progression. Data quality was ensured through a protocol that included verifying medications by cross-referencing FDA approval status \citep{Stocchi2024} and dosing guidelines using RxNorm codes and Athena. LEDD was calculated between consecutive visits using the standardized conversion factors established by \citep{Tomlinson2010} and updated in further studies \citep{Joshi2010, Kukkle2024}. The conversion factors proposed by \citep{Jost2023} were integrated in our LEDD computation pipeline since they provided the most comprehensive and up-to-date systematic review of LEDD calculations, incorporating data from 64 studies and providing evidence-based conversion factors for 31 different antiparkinsonian medications (e.g., 100mg levodopa = 133mg entacapone = 1mg pramipexole), accounting for both immediate and controlled-release formulations, as well as varying routes of administration. Percentage changes between visits were computed through sign-preserving log transformation followed by Winsorization at the 5th and 95th percentiles \citep{Grissom2000}. Figure~\ref{fig:normalized_pct_change} illustrates the normalized percentage of LEDD change. A key characteristic of our dataset is the zero-inflation pattern, with 75\% of patients maintaining stable LEDD values between visits. 


\subsection{Two-Stage Modeling Approach}

\begin{figure}[H]
   \centering 
   \includegraphics[width=4.5in]{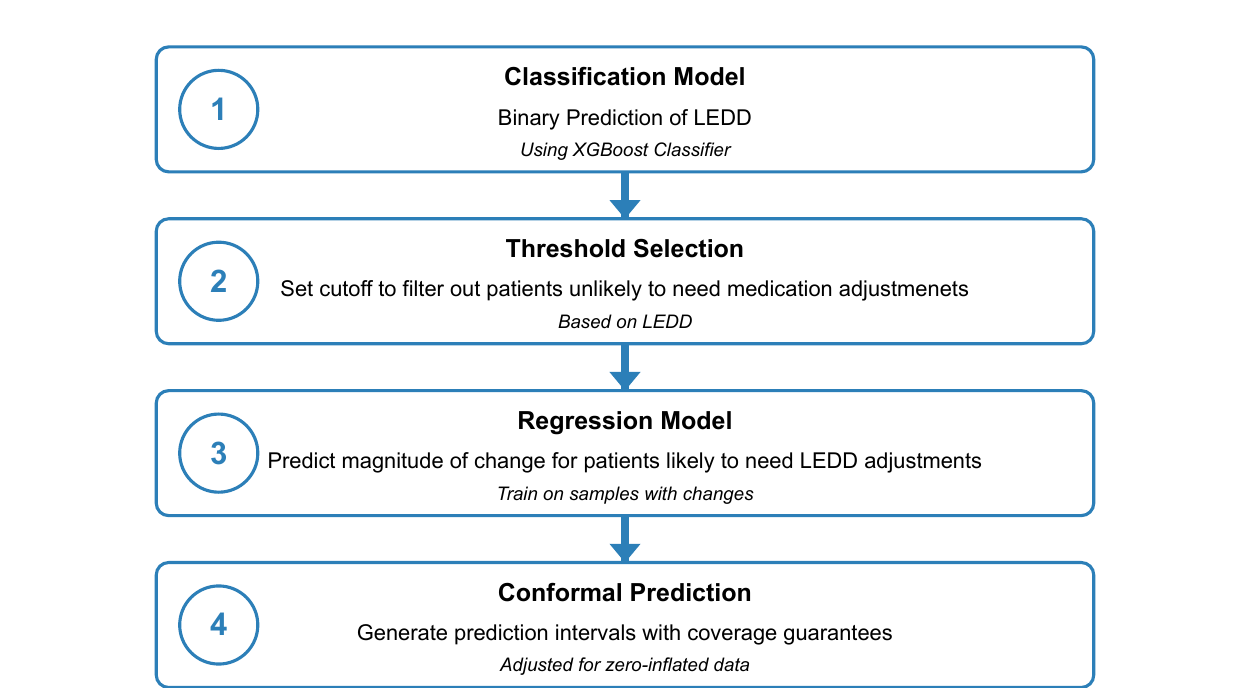} 
   \caption{Four-step workflow of our two-stage modeling approach: (1) classification of medication change likelihood, (2) threshold selection for patient filtering, (3) regression modeling for LEDD magnitude change, and (4) conformal prediction.}
   \label{fig:conformal_steps} 
 \end{figure}

Our approach addresses the zero-inflation challenge in PD medication prediction using a two-stage approach. This combines classification methods with conformal prediction to enhance statistical power while maintaining (asymptotic) coverage guarantees. Figure~\ref{fig:conformal_steps} illustrates the steps of our approach. 

Formally, let $Y$ represent the percentage change in LEDD, with many observations clustered at zero (unchanged medication), and let $X$ represent our feature vector. For PD medication data, we observe that:

\begin{equation}\label{eq:eq1}
    \mathbb{P}(Y=0 \mid X) > 0
\end{equation}

\paragraph {Step 1: Classification of Medication Change Likelihood.} We first train a classification model to estimate:   

\begin{equation}\label{eq:eq2}
    {p}(x) \;=\; \mathbb{P}\{Y \not= 0 \mid X = x\}.
\end{equation}

This represents the probability that a patient will require medication adjustment. For implementation, we use a gradient boosting classifier (XGBoost) \citep{Chen2016} that outputs $\widehat{p}(x)$, an estimate of this probability.

For a new patient with features $x_{\text{new}}$ we compute a classification cutoff $\alpha_{r}$ based on a hyperparameter $r\in[0,1]$
\begin{equation}\label{eq:eq3}
  \alpha_{r}
  \;=\;
  \mathrm{quantile}\Bigl(\{\widehat{p}(x_i) : (x_i,y_i)\in \mathcal{D}_{\text{cal,1}}\},\,r\Bigr),
\end{equation}
where $\mathcal{D}_{\text{cal},1}$ is a calibration dataset. This cutoff determines whether the patient is predicted to need medication adjustment.

\paragraph {Step 2: Conformal Prediction of LEDD Percentage Change}

For patients predicted to need adjustment where $\hat{p}(x_{new}) > \alpha_r$, we apply conformal prediction to construct an interval for the magnitude of change. Using a regression model $\hat{f}(x)$ trained only on samples with non-zero LEDD changes, we compute conformity scores:
\begin{equation}\label{eq:eq4}
S_i = |y_i - \hat{f}(x_i)|.
\end{equation}

For a desired coverage level $1 - \alpha$, we construct the prediction interval as:
\begin{equation}\label{eq:eq5}
C(x_{new}) = [\hat{f}(x_{new}) - q_\gamma, \hat{f}(x_{new}) + q_\gamma],
\end{equation}
where $q_\gamma$ is the appropriate quantile of conformity scores, adjusted to account for the classification step:
\begin{equation}\label{eq:eq6}
\gamma = \max\{0, \min\{1, \frac{1 - \hat{\beta} - r}{1 - r}\}\},
\end{equation}
where $\hat{\beta}$ estimates the proportion of true zero outcomes among those predicted as zero, i.e.,
\begin{equation}\label{eq:eq7}
\hat{\beta} = \frac{\sum_{(x_i, y_i) \in D_{val}} \mathbb{I}\{\hat{p}(x_i) \le \alpha_r \text{ and } y_i = 0\}}{\sum_{(x_i, y_i) \in D_{val}} \mathbb{I}\{\hat{p}(x_i) \le \alpha_r\}}.
\end{equation}

The final prediction interval for any new patient is:
\begin{equation}\label{eq:eq8}
  \widetilde{C}(x_{\text{new}}) 
  \;=\;
  \begin{cases}
    \{0\},
    \text{if}\;\widehat{p}(x_{\text{new}}) \;\le\; \alpha_{r}; \\
    \bigl[\widehat{f}(x_{\text{new}}) - q_{r},\widehat{f}(x_{\text{new}}) + q_{r}\bigr], 
    \text{otherwise.}
  \end{cases}
\end{equation}

We optimize the hyperparameter $r$ to achieve minimal average interval length while maintaining the specified coverage guarantee.

\subsection{Conformal Prediction Framework}
To quantify uncertainty in medication predictions, we implemented three established conformal prediction methods to provide statistically valid prediction intervals with guaranteed coverage. See Appendix~\ref{sec:cp_framework} for a comprehensive look. Each method presents distinct trade-offs between theoretical guarantees, computational efficiency, and empirical performance. CV+ balances statistical rigor with reasonable computational cost, J+aB excels in computational efficiency with ensemble methods, and Naïve prioritizes simplicity despite weaker guarantees. The three methods were integrated into our two-stage framework by first applying the classification model to identify patients likely to need medication adjustments, then constructing prediction intervals only for those patients using the selected conformal method.

\subsection{Evaluation Framework}

We evaluated our approach across several key metrics: First, the stage-one classification model was assessed using the area under the receiver operating characteristic curve (AUC), sensitivity, and specificity metrics through 10-fold cross-validation. For patients predicted to require medication adjustments, our second model evaluated regression accuracy using root mean squared error (RMSE), mean absolute error (MAE), and the coefficient of determination (R²). To assess the reliability of uncertainty estimates, we analyzed the performance of each conformal prediction method by examining empirical coverage (the proportion of true values falling within the prediction intervals), marginal coverage (average coverage across all cutoffs) and interval length. These evaluations were conducted at a nominal coverage of 80\%, which represents standard choice in conformal prediction literature \citep{Angelopoulos2023, Vovk2005} reflecting appropriate reliability-precision equilibrium \citep{Svensson2018, Papadopoulos2017, Papadopoulos2009} and allowing for a clinically meaningful balance between confidence and interval sharpness for decision support in high-stakes medication management scenarios. In addition, we trained separate models to forecast future medication needs via the LEDD and evaluated the accuracy and reliability of these forecasts across multiple time horizons (i.e., 6 months, 1 year, 2 years, and 4 years). Finally, to enhance interpretability and ensure clinical relevance, we employed Shapley value analysis \citep{Sundararajan2020} to identify the most influential features driving model predictions. This allowed us to better understand the key clinical factors associated with changes in medication regimens and support the potential integration of our model into real-world decision-making processes. 

\section{Results} 

\subsection{Performance of Classification and Regression Models}

Our two-stage approach leverages the complementary strengths of classification and regression modeling for LEDD prediction. We developed and trained the first-stage classification model to identify whether patients would experience LEDD changes (increase or decrease) across multiple time windows (i.e., 6 months, 1 year, 2 years, and 4 years). Using XGBoost with optimized hyperparameters (i.e., learning rate: 0.1, max depth: 7, number of boosting rounds: 700, alpha: 0.1, lambda: 1) via GridSearchCV and Bayesian Optimization \citep{Wu2019} along with L1/L2 regularization \citep{Ng2004} and early stopping, this model demonstrated strong prediction capabilities achieving an AUC of 0.98 for 6-month LEDD forecasting and maintaining robust performance (AUC $>$ 0.89) even when extended to 4-year prediction timeframes \citep{Ying2019}. Optimal configurations were consistent across time horizons. Appendix~\ref{sec:feature_importance} and Figure~\ref{fig:fig4} show mean feature importance across timeframes. 

For patients classified as likely to require medication adjustments, our second-stage regression model quantifies the expected percentage change in LEDD across patient visits.  




\subsection{Standard vs Two-Stage Conformal Prediction} 
\begin{figure}[t]
   \centering 
   \includegraphics[height=1.65in]{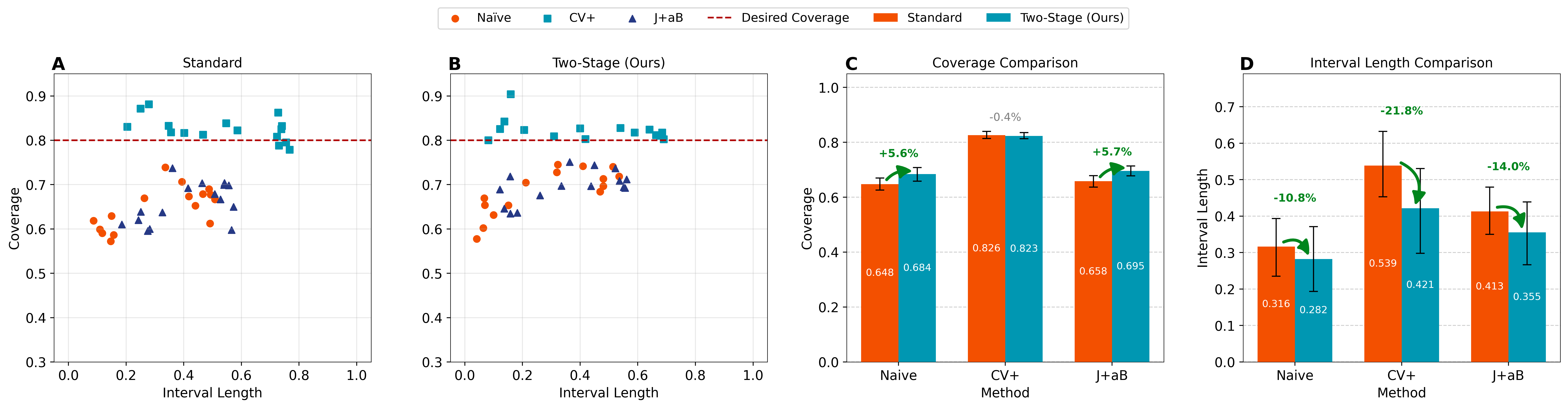} 
   \caption{(A-D) compare standard and our two-stage conformal approach. (A) and (B) show coverage vs. interval length at different classification cutoffs. (C-D) shows  comparison of marginal coverage (higher is better) and mean interval length (lower is better) with bootstrap confidence intervals for Naïve, CV+, and J+aB.}
   \label{fig:cp_comparison} 
 \end{figure}

Figure~\ref{fig:cp_comparison}A and Figure~\ref{fig:cp_comparison}B provide a side-by-side comparison between standard conformal prediction and our two-stage approach across different methods. Each point in the scatter plot represents performance at a specific cutoff, illustrating the relationship between coverage and interval length; the ideal performance combines high coverage with narrow intervals. While standard conformal methods struggle to achieve the desired 80\% nominal coverage (red dashed line), our two-step approach enables significantly more points reaching and exceeding it, particularly for CV+. Our two-stage-inspired CV+ method demonstrates notable improvement, maintaining equivalent marginal coverage while reducing interval lengths compared to standard. For Naïve and J+aB methods, our approach shows improvements in both metrics simultaneously, enhancing coverage while reducing interval length, as evidenced by the positioning of data points in Figure~\ref{fig:cp_comparison}A and Figure~\ref{fig:cp_comparison}B. Figure~\ref{fig:cp_comparison}C and Figure~\ref{fig:cp_comparison}D quantify these improvements by averaging performance across all classification cutoffs (0.0-0.95). Our two-stage CV+ method achieved a marginal coverage of 82.3\%, exceeding nominal coverage (80\%) while requiring 21.8\% shorter prediction intervals than standard CV+ (0.539 to 0.421). This slight over-coverage indicates appropriate conservatism in the prediction intervals, ensuring reliability while significantly improving precision for clinical applications. Although both standard and two-stage CV+ maintain nearly equivalent marginal coverage (82.6\% vs 82.3\%), the substantial decrease in interval length of our CV+ approach without sacrificing reliability represents a significant advancement for clinical decision-making. 
 
The Naïve method showed a 10.8\% reduction in interval length (0.316 to 0.282) while simultaneously increasing coverage by 5.6\% (64.8\% to 68.4\%), bringing it closer to the nominal target. Similarly, J+aB demonstrated a 14\% reduction in interval length (0.413 to 0.355) with a 5.7\% increase in coverage (65.8\% to 69.5\%). 

These results show that our two-stage approach consistently outperforms standard conformal prediction across all tested methods and metrics. To further validate the necessity of our approach, we compared against standard single-stage conformal prediction baselines that do not explicitly handle zero-inflation (Appendix~\ref{sec:baseline_comparison}) demonstrating substantial improvements with coverage gains up to 20.3 percentage points (53.6\% to 73.9\%) and interval length reductions up to 0.619 units (0.811 to 0.192). Additional statistical validation through paired t-tests and bootstrap confidence intervals confirms the significance of these improvements (Appendix~\ref{sec:experiments}). The CV+ method shows the largest reduction in interval length while maintaining comparable coverage, while both Naïve and J+aB exhibit improvements in both coverage and interval length. This consistent pattern of enhancement across different conformal methods suggests the robustness of our approach for handling zero-inflated PD medication data. To understand the underlying reasons for these performance differences, we next analyze the calibration properties of each method. 

\subsection{Uncertainty Quantification Across Time} 

\begin{figure}[t]
   \centering 
   \includegraphics[width=3.0in]{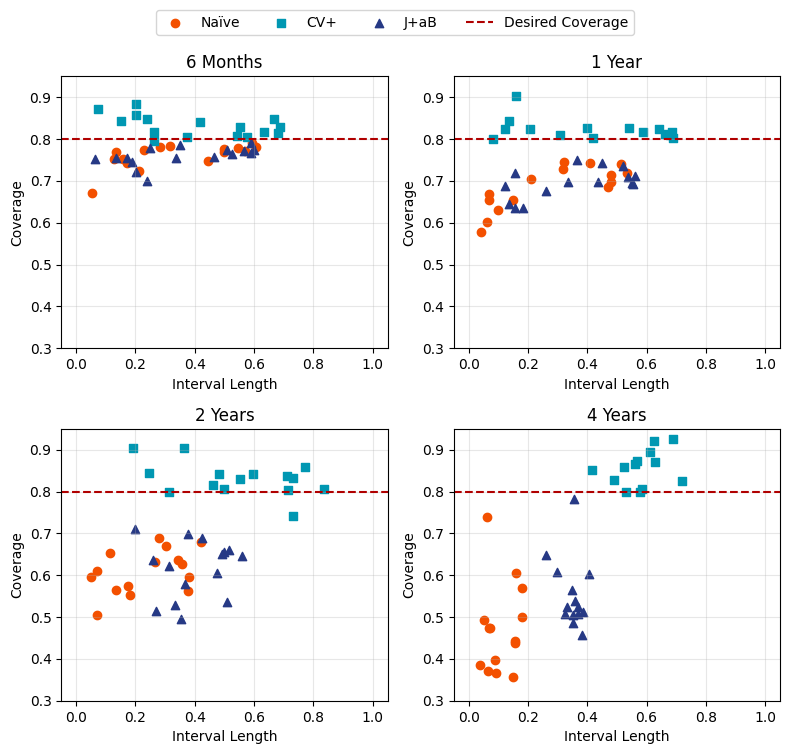} 
   \caption{Uncertainty quantification across time horizons (6 months, 1 year, 2 years, 4 years) for Naïve, CV+, and J+aB. Prediction uncertainty changes over progressively longer forecasting periods.}
   \label{fig:cp_through_time} 
 \end{figure}

We examined calibration quality by comparing nominal and empirical coverage. Detailed analysis (Appendix~\ref{sec:calib_analysis}) revealed that our two-stage CV+ approach demonstrates excellent calibration (mean error +2.3\%), while Naïve and J+aB methods showed substantial under-coverage despite interval length improvements. These calibration properties directly impact prediction reliability across different time horizons, as we examine next. Figure~\ref{fig:cp_through_time} illustrates how our approach performs across different prediction timeframes (i.e., 6 months, 1 year, 2 years, and 4 years).
For short-term predictions (6 months), our two-stage CV+ approach consistently exceeds nominal coverage for multiple cutoffs while maintaining competitive interval lengths. As the prediction horizon extends to 1 year, CV+ maintains strong coverage, while Naïve and J+aB show modest declines. For medium and long-term predictions (2-4 years), the performance differences between methods becomes more pronounced. CV+ maintains coverage above 80\% across most cutoffs for 2-year predictions, though with increased interval lengths. At 4 years, the Naïve method exhibits significant coverage fluctuations, while J+aB and CV+ show increasingly wide prediction intervals. 

 Table~\ref{tab:cp_metrics_table} presents the detailed performance of our approach across different time horizons and selected cutoffs. The RMSE at different cutoffs reveals a significant relationship between classifier confidence and regression accuracy. As the classification cutoffs increase, our model selectively focuses on patients with higher probability of medication changes, resulting in enhanced predictive precision. Bolded pairs of coverage and interval length represent the optimal trade-off between statistical reliability and prediction precision, highlighting the best-performing method for each specific timeframe. 
 

\begin{table}[t]
  \centering
  \small
  \caption{Performance metrics across time periods and cutoffs for three conformal methods. Optimal configurations are highlighted, showing trade-offs between coverage guarantees and prediction precision with extending time horizons.}
  \begin{tabular}{llc|cc|cc|cc}
    \toprule
    \textbf{Time} & \textbf{Cutoff} & \textbf{RMSE} 
    & \multicolumn{2}{c|}{\textbf{Naïve}} 
    & \multicolumn{2}{c|}{\textbf{CV+}} 
    & \multicolumn{2}{c}{\textbf{J+aB}} \\
    \cmidrule(lr){4-5} \cmidrule(lr){6-7} \cmidrule(l){8-9}
    & & & \textbf{Coverage} & \textbf{Length} 
        & \textbf{Coverage} & \textbf{Length} 
        & \textbf{Coverage} & \textbf{Length} \\
    \midrule
    \multirow{3}{*}{6M} 
      & 0    & 0.176 & 77.3\% & 0.230 & 79.5\% & 0.262 & 77.9\% & 0.248 \\
      & 0.5  & 0.142 & 78.0\% & 0.283 & 84.1\% & 0.418 & 75.5\% & 0.336 \\
      & 0.95 & 0.122 & 67.0\% & 0.052 & \textbf{87.3\%} & \textbf{0.075} & 75.1\% & 0.064 \\
    \midrule
    \multirow{3}{*}{1Y} 
      & 0    & 0.190 & 74.5\% & 0.322 & 80.3\% & 0.419 & 75.1\% & 0.364 \\
      & 0.5  & 0.149 & 70.5\% & 0.212 & 82.7\% & 0.400 & 69.7\% & 0.335 \\
      & 0.95 & 0.060 & 57.8\% & 0.041 & \textbf{80.0\%} & \textbf{0.082} & 68.9\% & 0.122 \\
    \midrule
    \multirow{3}{*}{2Y} 
      & 0    & 0.200 & 68.9\% & 0.280 & 80.6\% & 0.500 & 69.9\% & 0.379 \\
      & 0.5  & 0.286 & 55.3\% & 0.182 & 81.6\% & 0.462 & 57.9\% & 0.366 \\
      & 0.9  & 0.251 & 59.6\% & 0.051 & \textbf{90.4\%} & \textbf{0.192} & 71.2\% & 0.200 \\
    \midrule
    \multirow{3}{*}{4Y} 
      & 0    & 0.189 & 60.5\% & 0.161 & 85.3\% & 0.417 & 64.9\% & 0.262 \\
      & 0.5  & 0.265 & 47.4\% & 0.069 & 92.1\% & 0.626 & 56.6\% & 0.349 \\
      & 0.85 & 0.141 & \textbf{73.9\%} & \textbf{0.063} & 100\% & 0.685 & 78.3\% & 0.355 \\
    \bottomrule
  \end{tabular}
  \label{tab:cp_metrics_table}
\end{table}

Notably, at 6 months the optimal balance between coverage and interval length is achieved at cutoff 0.95, where CV+ reaches 87.3\% coverage with a narrow interval length of 0.075. This translates to predicting medication changes within ±3.75\% of the current dosage for 87.3\% of patients. While Naïve and J+aB methods achieve marginally narrower intervals (0.052 and 0.064 respectively), they do so at the cost of substantially reduced coverage (67\% and 75.1\%). For one-year predictions, CV+ maintained 80\% coverage with an interval length of 0.082 at the 0.95 cutoff, providing clinicians with reliable estimates for annual planning within ±4.1\% of current dosage. The performance extends to medium-term predictions (2 years), where CV+ achieved 90.4\% coverage with an interval length of 0.192 at cutoff of 0.9, translating to expected LEDD changes within ± 9.6\% of current dosage. During this same period, Naïve and J+aB showed reduced coverage (59.6\% and 71.2\% respectively).  

The 4-year prediction horizon reveals distinct pattern shifts in method performance. CV+ shows over-coverage at cutoff 0.85, achieving 100\% coverage with a conservative interval length of 0.685 (±34.25\% of current dosage). Despite its theoretical reliability and coverage guarantees, such wide intervals render CV+ less practical for clinical applications at this time period, as it provides limited practical guidance for long-term medication management. Interestingly, the Naïve method shows potential utility for 4-year predictions despite its variable coverage (from 35.7\% at cutoff 0.2 to 73.9\%). At cutoff 0.85, it achieves 73.9\% coverage with a notably narrow interval length of 0.063 (±3.15\% medication change), but clinicians should be aware that this approach may underestimate uncertainty in roughly 26.1\% of cases. The J+aB method provides moderate coverage (from 45.8\% at cutoff 0.4 to 78.3\%) with its best performance at cutoff 0.85 (78.3\% coverage, 0.355 interval length) translating into ±17.75\% medication changes. Like CV+, its intervals remain too wide to provide meaningful guidance compared to the Naïve method's more precise predictions. The trade-off between coverage and interval length becomes more pronounced over longer time horizons, with all methods requiring wider intervals to achieve acceptable coverage. This pattern reflects the increasing uncertainty inherent in long-term medication forecasting, with different methods offering varying balances between coverage guarantees and prediction precision as the timeframe extends. Overall, our findings further establish our two-stage approach as a reliable framework for uncertainty quantification in medication prediction, with different configurations offering optimal performance depending on the timeframe. 

\section{Discussion} 

\paragraph{Clinical Relevance}
Our approach demonstrates significant clinical utility across different PD medication management timeframes. Short-term predictions (87\% coverage at ±3.75\% LEDD change for 6 months), enable evidence-based dose adjustments during routine follow-up visits. This precision is valuable for patients with emerging motor fluctuations or early dyskinesia, where even small dosage adjustments can significantly impact symptom control.  

For 1-year treatment planning, we achieve nominal coverage with ±4.1\% LEDD changes, allowing neurologists to confidently plan medication adjustments across multiple quarterly visits while accounting for potential disease progression. This timeframe is particularly valuable for optimizing the balance between symptom control and minimizing the risk of motor complications during a critical period when many patients can experience fluctuating responses to levodopa, though this timing varies among individuals \citep{Espay2018}. Medium-term treatment predictions (±9.6\% LEDD changes at 2 years) enable proactive identification of patients likely to require substantial LEDD increases. This foresight allows earlier consideration of alternatives to oral medication management, including therapeutic interventions (physical therapy, occupational therapy, speech and swallowing therapy), device-based approaches (deep brain stimulation, duopa pump), and emerging non-invasive brain stimulation techniques (focused ultrasound, transcranial magnetic stimulation, transcranial direct current stimulation) \citep{madrid2021non}.  

Our prediction intervals might assist neurologists and movement disorder specialists to implement targeted monitoring protocols for patients at higher risk of developing levodopa-induced complications based on forecasted LEDD changes. This approach may be particularly beneficial for patient subpopulations with increased vulnerability to dyskinesia, including women, younger patients, and those with lower body weight \citep{sharma2010classifying}.

For long-term management (4+ years), despite inherently wider intervals reflecting growing uncertainty in medication needs, our approach can help clinicians identify potential medication trajectory patterns. These longer-term predictions can supplement, rather than replace, clinical judgment, allowing for informed discussions about long-term disease management strategies, helping set appropriate expectations with patients and caregivers about future medication needs. 

\paragraph{Theoretical and Empirical Insights from Conformal Approaches}
Our findings reveal important patterns that directly reflect the theoretical underpinnings of the conformal methods we evaluated. The superior performance of CV+ for short and medium-term predictions (82.3\% marginal coverage compared to 69.5\% for J+aB and 68.4\% for Naïve) validates its theoretical advantage of using multiple distinct calibration sets. While CV+ and J+aB share the same theoretical coverage guarantee (at least $1 - 2\alpha$), CV+'s empirical superiority demonstrates how implementation differences impact real-world performance. This stability is particularly valuable for capturing heterogeneity in PD increasing mediation needs. 

The reduction in interval length by 21.8\% of our two-stage conformal approach while maintaining robust coverage represents a clinically meaningful improvement in prediction precision, demonstrating how methodological innovations can address domain-specific challenges like zero-inflation. The calibration error landscape (Appendix~\ref{sec:calib_analysis}, Figure~\ref{fig:mean_calib_error}) further highlights this: standard CV+ shows a small positive error (0.026), indicating slight over-coverage which our framework mitigates by focusing on the most relevant prediction targets, while both Naïve and J+aB show substantial negative errors (-0.152 and -0.142 respectively), reflecting under-coverage consistent with their theoretical limitations: the Naïve method lacks distribution-free guarantees, while J+aB's performance is likely affected by the correlation between leave-one-out estimates in our dataset's size. 

The crossover in performance observed at longer time horizons, where Naïve methods begin to outperform CV+ for 4-year predictions, reveals an interesting methodological insight: as inherent uncertainty increases substantially, the theoretical advantages of more sophisticated methods may diminish. This suggests that method selection should be time-horizon dependent, an important consideration for longitudinal prediction tasks across various domains. Additionally, the consistent pattern across all methods showing wider intervals at longer time horizons reflects a fundamental statistical principle: uncertainty compounds over time. This methodological observation validates our approach's ability to quantify this increasing uncertainty appropriately, despite the challenging zero-inflated distribution in PD medication data. These methodological findings extend beyond our specific application to inform conformal prediction methodology more broadly, particularly for longitudinal medical predictions with class imbalance, highlighting how theoretical guarantees translate to empirical performance in challenging real-world datasets. 



\paragraph{Limitations and Future Work}

Despite encouraging results, our study has notable limitations. Our cutoff/threshold selection approach requires a more systematic methodology to optimally balance classification accuracy, regression performance, and interval length. Future work will explore automatic cutoff optimization and advanced conformal methods to address distribution shifts inherent in disease progression  (Appendix~\ref{sec:cp_distribution_shift}). We aim to incorporate outpatient and multi-center data, and develop multi-dimensional outcome predictions encompassing motor symptoms (UPDRS), non-motor symptoms (NMSS), and quality of life measures (PDQ-39). Additional enhancements include integrating inpatient risk factors to anticipate complications affecting medication decisions and leveraging large language models to extract subtle progression indicators from unstructured clinical narratives.

\paragraph{Code and Data Availability}

Code to reproduce all figures, statistical analyses, and results is available at \url{https://github.com/rdiazrincon/two-stage_conformal_pd}. The University of Florida Health dataset cannot be shared due to privacy regulations.

\bibliography{sample}

\begin{thebibliography}{66}
\providecommand{\natexlab}[1]{#1}
\providecommand{\url}[1]{\texttt{#1}}
\expandafter\ifx\csname urlstyle\endcsname\relax
  \providecommand{\doi}[1]{doi: #1}\else
  \providecommand{\doi}{doi: \begingroup \urlstyle{rm}\Url}\fi

\bibitem[Alaa and Schaar(2020)]{Alaa2020}
Ahmed Alaa and Mihaela Van~Der Schaar.
\newblock Discriminative jackknife: Quantifying uncertainty in deep learning via higher-order influence functions.
\newblock In \emph{International Conference on Machine Learning}, pages 165--174, 2020.

\bibitem[Alnemer et~al.(2016)Alnemer, Rajab, and Aljarah]{Alnemer2016}
Loai~M. Alnemer, Lama Rajab, and Ibrahim Aljarah.
\newblock Conformal prediction technique to predict breast cancer survivability.
\newblock \emph{International Journal of Advanced Science and Technology}, 96:\penalty0 1--10, 11 2016.
\newblock ISSN 20054238.
\newblock \doi{10.14257/ijast.2016.96.01}.

\bibitem[Aminoff et~al.(2011)Aminoff, Christine, Friedman, Chou, Lyons, Pahwa, Bloem, Parashos, Price, Malaty, Iansek, Bodis-Wollner, Suchowersky, Oertel, Zamudio, Oberdorf, Schmidt, and Okun]{Aminoff2011}
Michael~J. Aminoff, Chad~W. Christine, Joseph~H. Friedman, Kelvin~L. Chou, Kelly~E. Lyons, Rajesh Pahwa, Bastian~R. Bloem, Sotirios~A. Parashos, Catherine~C. Price, Irene~A. Malaty, Robert Iansek, Ivan Bodis-Wollner, Oksana Suchowersky, Wolfgang~H. Oertel, Jorge Zamudio, Joyce Oberdorf, Peter Schmidt, and Michael~S. Okun.
\newblock Management of the hospitalized patient with parkinson's disease: Current state of the field and need for guidelines.
\newblock \emph{Parkinsonism and Related Disorders}, 17:\penalty0 139--145, 3 2011.
\newblock ISSN 13538020.
\newblock \doi{10.1016/j.parkreldis.2010.11.009}.

\bibitem[Angelopoulos et~al.(2020)Angelopoulos, Bates, Malik, and Jordan]{Angelopoulos2020}
Anastasios Angelopoulos, Stephen Bates, Jitendra Malik, and Michael~I. Jordan.
\newblock Uncertainty sets for image classifiers using conformal prediction.
\newblock \emph{arxiv preprint}, 9 2020.
\newblock URL \url{http://arxiv.org/abs/2009.14193}.

\bibitem[Angelopoulos and Bates(2021)]{Angelopoulos2021}
Anastasios~N. Angelopoulos and Stephen Bates.
\newblock A gentle introduction to conformal prediction and distribution-free uncertainty quantification.
\newblock \emph{arXiv preprint}, 7 2021.
\newblock URL \url{http://arxiv.org/abs/2107.07511}.

\bibitem[Angelopoulos and Bates(2023)]{Angelopoulos2023}
Anastasios~N. Angelopoulos and Stephen Bates.
\newblock Conformal prediction: A gentle introduction.
\newblock \emph{Foundations and Trends® in Machine Learning}, 16:\penalty0 494--591, 2023.
\newblock ISSN 1935-8237.
\newblock \doi{10.1561/2200000101}.

\bibitem[Angelopoulos et~al.(2024)Angelopoulos, Barber, and Bates]{Angelopoulos2024}
Anastasios~N Angelopoulos, Rina~Foygel Barber, and Stephen Bates.
\newblock Theoretical foundations of conformal prediction.
\newblock \emph{arXiv preprint}, 2024.

\bibitem[Armstrong and Okun(2020)]{Armstrong2020}
Melissa~J. Armstrong and Michael~S. Okun.
\newblock Diagnosis and treatment of parkinson disease: A review.
\newblock \emph{JAMA - Journal of the American Medical Association}, 323:\penalty0 548--560, 2 2020.
\newblock ISSN 15383598.
\newblock \doi{10.1001/jama.2019.22360}.

\bibitem[Barber et~al.(2021)Barber, Candes, Ramdas, and Tibshirani]{Barber2021}
Rina~Foygel Barber, Emmanuel~J Candes, Aaditya Ramdas, and Ryan~J Tibshirani.
\newblock Predictive inference with the jackknife+.
\newblock \emph{The Annals of Statistics}, 49:\penalty0 486--507, 2021.

\bibitem[Begoli et~al.(2019)Begoli, Bhattacharya, and Kusnezov]{Begoli2019}
Edmon Begoli, Tanmoy Bhattacharya, and Dimitri Kusnezov.
\newblock The need for uncertainty quantification in machine-assisted medical decision making, 1 2019.
\newblock ISSN 25225839.

\bibitem[Belyaev et~al.(2023)Belyaev, Murugappan, Velichko, and Korzun]{Belyaev2023}
Maksim Belyaev, Murugappan Murugappan, Andrei Velichko, and Dmitry Korzun.
\newblock Entropy-based machine learning model for fast diagnosis and monitoring of parkinson's disease.
\newblock \emph{Sensors (Basel, Switzerland)}, 23, 10 2023.
\newblock ISSN 14248220.
\newblock \doi{10.3390/s23208609}.

\bibitem[Chen and Guestrin(2016)]{Chen2016}
Tianqi Chen and Carlos Guestrin.
\newblock Xgboost: A scalable tree boosting system.
\newblock In \emph{Proceedings of the ACM SIGKDD International Conference on Knowledge Discovery and Data Mining}, volume 13-17-August-2016, pages 785--794. Association for Computing Machinery, 8 2016.
\newblock ISBN 9781450342322.
\newblock \doi{10.1145/2939672.2939785}.

\bibitem[Chernozhukov et~al.(2021)Chernozhukov, Wüthrich, and Zhu]{Chernozhukov2021}
Victor Chernozhukov, Kaspar Wüthrich, and Yinchu Zhu.
\newblock Distributional conformal prediction.
\newblock \emph{Proceedings of the National Academy of Sciences}, 118:\penalty0 e2107794118, 2021.

\bibitem[Cotzias et~al.(1967)Cotzias, Woert, and Schiffer]{Cotzias1967}
George~C Cotzias, Melvin H~Van Woert, and Lewis~M Schiffer.
\newblock Aromatic amino acids and modification of parkinsonism.
\newblock \emph{New England Journal of Medicine}, 276:\penalty0 374--379, 1967.

\bibitem[Devetyarov et~al.(2012)Devetyarov, Nouretdinov, Burford, Camuzeaux, Gentry-Maharaj, Tiss, Smith, Luo, Chervonenkis, Hallett, Vovk, Waterfield, Cramer, Timms, Sinclair, Menon, Jacobs, and Gammerman]{Devetyarov2012}
Dmitry Devetyarov, Ilia Nouretdinov, Brian Burford, Stephane Camuzeaux, Aleksandra Gentry-Maharaj, Ali Tiss, Celia Smith, Zhiyuan Luo, Alexey Chervonenkis, Rachel Hallett, Volodya Vovk, Mike Waterfield, Rainer Cramer, John~F. Timms, John Sinclair, Usha Menon, Ian Jacobs, and Alex Gammerman.
\newblock Conformal predictors in early diagnostics of ovarian and breast cancers.
\newblock \emph{Progress in Artificial Intelligence}, 1:\penalty0 245--257, 9 2012.
\newblock ISSN 21926360.
\newblock \doi{10.1007/s13748-012-0021-y}.

\bibitem[Dorsey et~al.(2020)Dorsey, Omberg, Waddell, Adams, Adams, Ali, Amodeo, Arky, Augustine, DInesh, Hoque, Glidden, Jensen-Roberts, Kabelac, Katabi, Kieburtz, Kinel, Little, Lizarraga, Myers, Riggare, Rosero, Saria, Schifitto, Schneider, Sharma, Shoulson, Stevenson, Tarolli, Luo, and McDermott]{Dorsey2020}
E.~Ray Dorsey, Larsson Omberg, Emma Waddell, Jamie~L. Adams, Roy Adams, Mohammad~Rafayet Ali, Katherine Amodeo, Abigail Arky, Erika~F. Augustine, Karthik DInesh, Mohammed~Ehsan Hoque, Alistair~M. Glidden, Stella Jensen-Roberts, Zachary Kabelac, DIna Katabi, Karl Kieburtz, Daniel~R. Kinel, Max~A. Little, Karlo~J. Lizarraga, Taylor Myers, Sara Riggare, Spencer~Z. Rosero, Suchi Saria, Giovanni Schifitto, Ruth~B. Schneider, Gaurav Sharma, Ira Shoulson, E.~Anna Stevenson, Christopher~G. Tarolli, Jiebo Luo, and Michael~P. McDermott.
\newblock Deep phenotyping of parkinson's disease, 2020.
\newblock ISSN 1877718X.

\bibitem[Espay et~al.(2018)Espay, Morgante, Merola, Fasano, Marsili, Fox, Bezard, Picconi, Calabresi, and Lang]{Espay2018}
Alberto~J. Espay, Francesca Morgante, Aristide Merola, Alfonso Fasano, Luca Marsili, Susan~H. Fox, Erwan Bezard, Barbara Picconi, Paolo Calabresi, and Anthony~E. Lang.
\newblock Levodopa-induced dyskinesia in parkinson disease: Current and evolving concepts, 12 2018.
\newblock ISSN 15318249.

\bibitem[Fannjiang et~al.(2022)Fannjiang, Bates, Angelopoulos, Listgarten, and Jordan]{Fannjiang2022}
Clara Fannjiang, Stephen Bates, Anastasios~N Angelopoulos, Jennifer Listgarten, and Michael~I Jordan.
\newblock Conformal prediction under feedback covariate shift for biomolecular design.
\newblock \emph{Proceedings of the National Academy of Sciences}, 119:\penalty0 e2204569119, 2022.

\bibitem[Fereshtehnejad et~al.(2015)Fereshtehnejad, Romenets, Anang, Latreille, Gagnon, and Postuma]{Fereshtehnejad2015}
Seyed~Mohammad Fereshtehnejad, Silvia~Rios Romenets, Julius~B.M. Anang, Véronique Latreille, Jean~François Gagnon, and Ronald~B. Postuma.
\newblock New clinical subtypes of parkinson disease and their longitudinal progression a prospective cohort comparison with other phenotypes.
\newblock \emph{JAMA Neurology}, 72:\penalty0 863--873, 8 2015.
\newblock ISSN 21686149.
\newblock \doi{10.1001/jamaneurol.2015.0703}.

\bibitem[Fox et~al.(2018)Fox, Katzenschlager, Lim, Barton, de~Bie, Seppi, Coelho, and Sampaio]{Fox2018}
Susan~H. Fox, Regina Katzenschlager, Shen~Yang Lim, Brandon Barton, Rob~M.A. de~Bie, Klaus Seppi, Miguel Coelho, and Cristina Sampaio.
\newblock International parkinson and movement disorder society evidence-based medicine review: Update on treatments for the motor symptoms of parkinson's disease, 8 2018.
\newblock ISSN 15318257.

\bibitem[Gerlach et~al.(2011)Gerlach, Winogrodzka, and Weber]{Gerlach2011}
Oliver~H.H. Gerlach, Ania Winogrodzka, and Wim~E.J. Weber.
\newblock Clinical problems in the hospitalized parkinson's disease patient: Systematic review, 2 2011.
\newblock ISSN 08853185.

\bibitem[Goetz(2003)]{Goetz2003}
Christopher~C. Goetz.
\newblock The unified parkinson's disease rating scale (updrs): Status and recommendations, 7 2003.
\newblock ISSN 08853185.

\bibitem[Goetz et~al.(1988)Goetz, Tanner, Stebbins, and Buchman]{goetz1988risk}
Christopher~G Goetz, Caroline~M Tanner, Glenn~T Stebbins, and Aron~S Buchman.
\newblock Risk factors for progression in parkinson's disease.
\newblock \emph{Neurology}, 38\penalty0 (12):\penalty0 1841--1841, 1988.

\bibitem[Goetz et~al.(2008)Goetz, Tilley, Shaftman, Stebbins, Fahn, Martinez-Martin, Poewe, Sampaio, Stern, Dodel, Dubois, Holloway, Jankovic, Kulisevsky, Lang, Lees, Leurgans, LeWitt, Nyenhuis, Olanow, Rascol, Schrag, Teresi, van Hilten, LaPelle, Agarwal, Athar, Bordelan, Bronte-Stewart, Camicioli, Chou, Cole, Dalvi, Delgado, Diamond, Dick, Duda, Elble, Evans, Evidente, Fernandez, Fox, Friedman, Fross, Gallagher, Goetz, Hall, Hermanowicz, Hinson, Horn, Hurtig, Kang, Kleiner-Fisman, Klepitskaya, Kompoliti, Lai, Leehey, Leroi, Lyons, McClain, Metzer, Miyasaki, Morgan, Nance, Nemeth, Pahwa, Parashos, Schneider, Sethi, Shulman, Siderowf, Silverdale, Simuni, Stacy, Stern, Stewart, Sullivan, Swope, Wadia, Walker, Walker, Weiner, Wiener, Wilkinson, Wojcieszek, Wolfrath, Wooten, Wu, Zesiewicz, and Zweig]{Goetz2008}
Christopher~G. Goetz, Barbara~C. Tilley, Stephanie~R. Shaftman, Glenn~T. Stebbins, Stanley Fahn, Pablo Martinez-Martin, Werner Poewe, Cristina Sampaio, Matthew~B. Stern, Richard Dodel, Bruno Dubois, Robert Holloway, Joseph Jankovic, Jaime Kulisevsky, Anthony~E. Lang, Andrew Lees, Sue Leurgans, Peter~A. LeWitt, David Nyenhuis, C.~Warren Olanow, Olivier Rascol, Anette Schrag, Jeanne~A. Teresi, Jacobus~J. van Hilten, Nancy LaPelle, Pinky Agarwal, Saima Athar, Yvette Bordelan, Helen~M. Bronte-Stewart, Richard Camicioli, Kelvin Chou, Wendy Cole, Arif Dalvi, Holly Delgado, Alan Diamond, Jeremy~P. Dick, John Duda, Rodger~J. Elble, Carol Evans, Virgilio~G. Evidente, Hubert~H. Fernandez, Susan Fox, Joseph~H. Friedman, Robin~D. Fross, David Gallagher, Christopher~G. Goetz, Deborah Hall, Neal Hermanowicz, Vanessa Hinson, Stacy Horn, Howard Hurtig, Un~Jung Kang, Galit Kleiner-Fisman, Olga Klepitskaya, Katie Kompoliti, Eugene~C. Lai, Maureen~L. Leehey, Iracema Leroi, Kelly~E. Lyons, Terry McClain, Steven~W. Metzer, Janis
  Miyasaki, John~C. Morgan, Martha Nance, Joanne Nemeth, Rajesh Pahwa, Sotirios~A. Parashos, Jay~S. Schneider, Kapil Sethi, Lisa~M. Shulman, Andrew Siderowf, Monty Silverdale, Tanya Simuni, Mark Stacy, Matthew~B. Stern, Robert~Malcolm Stewart, Kelly Sullivan, David~M. Swope, Pettaruse~M. Wadia, Richard~W. Walker, Ruth Walker, William~J. Weiner, Jill Wiener, Jayne Wilkinson, Joanna~M. Wojcieszek, Summer Wolfrath, Frederick Wooten, Allen Wu, Theresa~A. Zesiewicz, and Richard~M. Zweig.
\newblock Movement disorder society-sponsored revision of the unified parkinson's disease rating scale (mds-updrs): Scale presentation and clinimetric testing results.
\newblock \emph{Movement Disorders}, 23:\penalty0 2129--2170, 11 2008.
\newblock ISSN 15318257.
\newblock \doi{10.1002/mds.22340}.

\bibitem[Grissom(2000)]{Grissom2000}
Robert~J Grissom.
\newblock Heterogeneity of variance in clinical data.
\newblock \emph{Journal of Consulting and Clinical Psychology}, 68:\penalty0 155--165, 2000.
\newblock \doi{10.1037/00}.

\bibitem[Gutowski et~al.(2023)Gutowski, Antkiewicz, and Szlufik]{Gutowski2023}
Tomasz Gutowski, Ryszard Antkiewicz, and Stanisław Szlufik.
\newblock Machine learning with optimization to create medicine intake schedules for parkinson’s disease patients.
\newblock \emph{PLoS ONE}, 18, 10 2023.
\newblock ISSN 19326203.
\newblock \doi{10.1371/journal.pone.0293123}.

\bibitem[Jankovic et~al.(1990)Jankovic, Mcdermott, Carter, Gauthier, Goetz, Golbe, Huber, Koller, Olanow, Shoulson, Stern, Tanner, Weiner, and Group]{Jankovic1990}
J~Jankovic, ;~M Mcdermott, ;~J Carter, ;~S Gauthier, ;~C Goetz, ;~L Golbe, ;~S Huber, ;~W Koller, ;~C Olanow, ;~I Shoulson, ;~M Stern, ;~C Tanner, ;~W Weiner, and Parkinson~Study Group.
\newblock Variable expression of parkinson's disease: A base-line analysis of the datatop cohort.
\newblock \emph{Neurology}, 1990.
\newblock URL \url{https://www.neurology.org}.

\bibitem[Jatain et~al.(2021)Jatain, Bajaj, Agarwal, and Bullah]{Jatain2021}
Aman Jatain, Shalini~Bhaskar Bajaj, Ritika Agarwal, and Haziq~Rahat Bullah.
\newblock Classification based levodopamine response prediction in parkinson’s disorder.
\newblock \emph{Applied Artificial Intelligence}, 35:\penalty0 1287--1303, 2021.
\newblock ISSN 10876545.
\newblock \doi{10.1080/08839514.2021.1975881}.

\bibitem[Jenner(2008)]{Jenner2008}
Peter Jenner.
\newblock Molecular mechanisms of l-dopa-induced dyskinesia, 9 2008.
\newblock ISSN 1471003X.

\bibitem[Joshi et~al.(2010)Joshi, Shenoy, Simha, Rrashmi, Venugopal, and Patnaik]{Joshi2010}
Sandhya Joshi, Deepa Shenoy, G.~G.~Vibhudendra Simha, P.~L. Rrashmi, K.~R. Venugopal, and L.~M. Patnaik.
\newblock Classification of alzheimer's disease and parkinson's disease by using machine learning and neural network methods.
\newblock In \emph{ICMLC 2010 - The 2nd International Conference on Machine Learning and Computing}, pages 218--222, 2010.
\newblock ISBN 9780769539775.
\newblock \doi{10.1109/ICMLC.2010.45}.

\bibitem[Jost et~al.(2023)Jost, Kaldenbach, Antonini, Martinez-Martin, Timmermann, Odin, Katzenschlager, Borgohain, Fasano, Stocchi, Hattori, Kukkle, Rodríguez-Violante, Falup-Pecurariu, Schade, Petry-Schmelzer, Metta, Weintraub, Deuschl, Espay, Tan, Bhidayasiri, Fung, Cardoso, Trenkwalder, Jenner, Chaudhuri, and Dafsari]{Jost2023}
Stefanie~T. Jost, Marie~Ann Kaldenbach, Angelo Antonini, Pablo Martinez-Martin, Lars Timmermann, Per Odin, Regina Katzenschlager, Rupam Borgohain, Alfonso Fasano, Fabrizio Stocchi, Nobutaka Hattori, Prashanth~Lingappa Kukkle, Mayela Rodríguez-Violante, Cristian Falup-Pecurariu, Sebastian Schade, Jan~Niklas Petry-Schmelzer, Vinod Metta, Daniel Weintraub, Guenther Deuschl, Alberto~J. Espay, Eng~King Tan, Roongroj Bhidayasiri, Victor~S.C. Fung, Francisco Cardoso, Claudia Trenkwalder, Peter Jenner, K.~Ray Chaudhuri, and Haidar~S. Dafsari.
\newblock Levodopa dose equivalency in parkinson's disease: Updated systematic review and proposals.
\newblock \emph{Movement Disorders}, 38:\penalty0 1236--1252, 7 2023.
\newblock ISSN 15318257.
\newblock \doi{10.1002/mds.29410}.

\bibitem[Kim et~al.(2020)Kim, Xu, and Barber]{Kim2020}
Byol Kim, Chen Xu, and Rina Barber.
\newblock Predictive inference is free with the jackknife+-after-bootstrap.
\newblock \emph{Advances in Neural Information Processing Systems}, 33:\penalty0 4138--4149, 2020.

\bibitem[Kim et~al.(2021)Kim, Suescun, Schiess, and Jiang]{Kim2021}
Yejin Kim, Jessika Suescun, Mya~C. Schiess, and Xiaoqian Jiang.
\newblock Computational medication regimen for parkinson’s disease using reinforcement learning.
\newblock \emph{Scientific Reports}, 11, 12 2021.
\newblock ISSN 20452322.
\newblock \doi{10.1038/s41598-021-88619-4}.

\bibitem[Kukkle et~al.(2024)Kukkle, Kalia, Habib, Jagota, Ojha, Kandadai, Desai, Caldera, Sirisena, Garg, Mestre, Neupane, Maytharakcheep, Sanyawut, and Borgohain]{Kukkle2024}
Prashanth~Lingappa Kukkle, Lorraine~V. Kalia, Ahsan Habib, Priya Jagota, Rajeev Ojha, Rukmini~Mridula Kandadai, Soaham Desai, Manjula Caldera, Darshana Sirisena, Divyani Garg, Tiago~A. Mestre, Rosy Neupane, Suppata Maytharakcheep, Kanyawat Sanyawut, and Rupam Borgohain.
\newblock Levodopa equivalent daily dosage: Geographical variations and real-life modules in parkinson's disease.
\newblock \emph{Movement Disorders Clinical Practice}, 11 2024.
\newblock ISSN 23301619.
\newblock \doi{10.1002/mdc3.14200}.

\bibitem[Latourelle et~al.(2017)Latourelle, Beste, Hadzi, Miller, Oppenheim, Valko, Wuest, Church, Khalil, Hayete, and Venuto]{Latourelle2017}
Jeanne~C. Latourelle, Michael~T. Beste, Tiffany~C. Hadzi, Robert~E. Miller, Jacob~N. Oppenheim, Matthew~P. Valko, Diane~M. Wuest, Bruce~W. Church, Iya~G. Khalil, Boris Hayete, and Charles~S. Venuto.
\newblock Large-scale identification of clinical and genetic predictors of motor progression in patients with newly diagnosed parkinson's disease: a longitudinal cohort study and validation.
\newblock \emph{The Lancet Neurology}, 16:\penalty0 908--916, 11 2017.
\newblock ISSN 14744465.
\newblock \doi{10.1016/S1474-4422(17)30328-9}.

\bibitem[Lusk et~al.(2023)Lusk, Choi, Clark, Johnson, Ford, Greiner, Goetz, Kaufman, O’Brien, and O’Brien]{Lusk2023}
Jay~B. Lusk, Sujung Choi, Amy~G. Clark, Kim Johnson, Cassie~B. Ford, Melissa~A. Greiner, Margarethe Goetz, Brystana~G. Kaufman, Richard O’Brien, and Emily~C. O’Brien.
\newblock Dementia and parkinson’s disease diagnoses in electronic health records vs. medicare claims data: a study of 101,980 linked patients.
\newblock \emph{BMC Neurology}, 23, 12 2023.
\newblock ISSN 14712377.
\newblock \doi{10.1186/s12883-023-03361-w}.

\bibitem[Madrid and Benninger(2021)]{madrid2021non}
Julian Madrid and David~H Benninger.
\newblock Non-invasive brain stimulation for parkinson’s disease: Clinical evidence, latest concepts and future goals: A systematic review.
\newblock \emph{Journal of neuroscience methods}, 347:\penalty0 108957, 2021.

\bibitem[Marek et~al.(2011)Marek, Jennings, Lasch, Siderowf, Tanner, Simuni, Coffey, Kieburtz, Flagg, Chowdhury, Poewe, Mollenhauer, Sherer, Frasier, Meunier, Rudolph, Casaceli, Seibyl, Mendick, Schuff, Zhang, Toga, Crawford, Ansbach, de~Blasio, Piovella, Trojanowski, Shaw, Singleton, Hawkins, Eberling, Russell, Leary, Factor, Sommerfeld, Hogarth, Pighetti, Williams, Standaert, Guthrie, Hauser, Delgado, Jankovic, Hunter, Stern, Tran, Leverenz, Baca, Frank, Thomas, Richard, Deeley, Rees, Sprenger, Lang, Shill, Obradov, Fernandez, Winters, Berg, Gauss, Galasko, Fontaine, Mari, Gerstenhaber, Brooks, Malloy, Barone, Longo, Comery, Ravina, Grachev, Gallagher, Collins, Widnell, Ostrowizki, Fontoura, La-Roche, Ho, Luthman, van~der Brug, Reith, and Taylor]{Marek2011}
Kenneth Marek, Danna Jennings, Shirley Lasch, Andrew Siderowf, Caroline Tanner, Tanya Simuni, Chris Coffey, Karl Kieburtz, Emily Flagg, Sohini Chowdhury, Werner Poewe, Brit Mollenhauer, Todd Sherer, Mark Frasier, Claire Meunier, Alice Rudolph, Cindy Casaceli, John Seibyl, Susan Mendick, Norbert Schuff, Ying Zhang, Arthur Toga, Karen Crawford, Alison Ansbach, Pasquale de~Blasio, Michele Piovella, John Trojanowski, Les Shaw, Andrew Singleton, Keith Hawkins, Jamie Eberling, David Russell, Laura Leary, Stewart Factor, Barbara Sommerfeld, Penelope Hogarth, Emily Pighetti, Karen Williams, David Standaert, Stephanie Guthrie, Robert Hauser, Holly Delgado, Joseph Jankovic, Christine Hunter, Matthew Stern, Baochan Tran, Jim Leverenz, Marne Baca, Sam Frank, Cathi~Ann Thomas, Irene Richard, Cheryl Deeley, Linda Rees, Fabienne Sprenger, Elisabeth Lang, Holly Shill, Sanja Obradov, Hubert Fernandez, Adrienna Winters, Daniela Berg, Katharina Gauss, Douglas Galasko, Deborah Fontaine, Zoltan Mari, Melissa Gerstenhaber, David
  Brooks, Sophie Malloy, Paolo Barone, Katia Longo, Tom Comery, Bernard Ravina, Igor Grachev, Kim Gallagher, Michelle Collins, Katherine~L. Widnell, Suzanne Ostrowizki, Paulo Fontoura, F.~Hoffmann La-Roche, Tony Ho, Johan Luthman, Marcel van~der Brug, Alastair~D. Reith, and Peggy Taylor.
\newblock The parkinson progression marker initiative (ppmi), 12 2011.
\newblock ISSN 03010082.

\bibitem[Ng(2004)]{Ng2004}
Andrew~Y Ng.
\newblock Feature selection, l 1 vs. l 2 regularization, and rotational invariance.
\newblock Technical report, 2004.

\bibitem[Olanow et~al.(2006)Olanow, Obeso, and Stocchi]{Olanow2006}
C~Warren Olanow, Jose~A Obeso, and Fabrizio Stocchi.
\newblock Continuous dopamine-receptor treatment of parkinson's disease: scientific rationale and clinical implications.
\newblock \emph{The Lancet Neurology}, 5:\penalty0 677--687, 2006.

\bibitem[Palacios et~al.(2011)Palacios, Gao, Mccullough, Jacobs, Patel, Mayo, Schwarzschild, and Ascherio]{Palacios2011}
Natalia Palacios, Xiang Gao, Marjorie~L. Mccullough, Eric~J. Jacobs, Alpa~V. Patel, Tinisha Mayo, Michael~A. Schwarzschild, and Alberto Ascherio.
\newblock Obesity, diabetes, and risk of parkinson's disease.
\newblock \emph{Movement Disorders}, 26:\penalty0 2253--2259, 10 2011.
\newblock ISSN 08853185.
\newblock \doi{10.1002/mds.23855}.

\bibitem[Papadopoulos et~al.(2002)Papadopoulos, Proedrou, Vovk, and Gammerman]{Papadopoulos2002}
Harris Papadopoulos, Kostas Proedrou, Volodya Vovk, and Alex Gammerman.
\newblock Inductive confidence machines for regression.
\newblock In \emph{Machine learning: ECML 2002: 13th European conference on machine learning Helsinki, Finland, August 19–23, 2002 proceedings 13}, pages 345--356, 2002.

\bibitem[Papadopoulos et~al.(2009{\natexlab{a}})Papadopoulos, Gammerman, and Vovk]{Papadopoulos2009}
Harris Papadopoulos, Alex Gammerman, and Volodya Vovk.
\newblock Reliable diagnosis of acute abdominal pain with conformal prediction.
\newblock \emph{Engineering Intelligent Systems}, 17:\penalty0 127, 2009{\natexlab{a}}.

\bibitem[Papadopoulos et~al.(2009{\natexlab{b}})Papadopoulos, Gammerman, and Vovk]{Papadopoulos2009b}
Harris Papadopoulos, Alex Gammerman, and Volodya Vovk.
\newblock Confidence predictions for the diagnosis of acute abdominal pain.
\newblock In \emph{IFIP international conference on artificial intelligence applications and innovations}, pages 175--184, 2009{\natexlab{b}}.

\bibitem[Papadopoulos et~al.(2017)Papadopoulos, Kyriacou, and Nicolaides]{Papadopoulos2017}
Harris Papadopoulos, Efthyvoulos Kyriacou, and Andrew Nicolaides.
\newblock Unbiased confidence measures for stroke risk estimation based on ultrasound carotid image analysis.
\newblock \emph{Neural Computing and Applications}, 28:\penalty0 1209--1223, 6 2017.
\newblock ISSN 09410643.
\newblock \doi{10.1007/s00521-016-2590-3}.

\bibitem[Postuma et~al.(2015)Postuma, Berg, Stern, Poewe, Olanow, Oertel, Obeso, Marek, Litvan, Lang, Halliday, Goetz, Gasser, Dubois, Chan, Bloem, Adler, and Deuschl]{Postuma2015}
Ronald~B. Postuma, Daniela Berg, Matthew Stern, Werner Poewe, C.~Warren Olanow, Wolfgang Oertel, José Obeso, Kenneth Marek, Irene Litvan, Anthony~E. Lang, Glenda Halliday, Christopher~G. Goetz, Thomas Gasser, Bruno Dubois, Piu Chan, Bastiaan~R. Bloem, Charles~H. Adler, and Günther Deuschl.
\newblock Mds clinical diagnostic criteria for parkinson's disease.
\newblock \emph{Movement Disorders}, 30:\penalty0 1591--1601, 10 2015.
\newblock ISSN 15318257.
\newblock \doi{10.1002/mds.26424}.

\bibitem[Prinster et~al.(2023)Prinster, Saria, and Liu]{Prinster2023}
Drew Prinster, Suchi Saria, and Anqi Liu.
\newblock Jaws-x: addressing efficiency bottlenecks of conformal prediction under standard and feedback covariate shift.
\newblock In \emph{International Conference on Machine Learning}, pages 28167--28190, 2023.

\bibitem[Prinster et~al.(2024)Prinster, Stanton, Liu, and Saria]{Prinster2024}
Drew Prinster, Samuel Stanton, Anqi Liu, and Suchi Saria.
\newblock Conformal validity guarantees exist for any data distribution (and how to find them).
\newblock \emph{arXiv preprint}, 2024.

\bibitem[Riasi et~al.(2024)Riasi, Delrobaei, and Salari]{Riasi2024}
Atiye Riasi, Mehdi Delrobaei, and Mehri Salari.
\newblock A decision support system based on recurrent neural networks to predict medication dosage for patients with parkinson's disease.
\newblock \emph{Scientific Reports}, 14, 12 2024.
\newblock ISSN 20452322.
\newblock \doi{10.1038/s41598-024-59179-0}.

\bibitem[Romano et~al.(2019)Romano, Patterson, and Candes]{Romano2019}
Yaniv Romano, Evan Patterson, and Emmanuel Candes.
\newblock Conformalized quantile regression.
\newblock In H~Wallach, H~Larochelle, A~Beygelzimer, F~d~Alché-Buc, E~Fox, and R~Garnett, editors, \emph{Advances in Neural Information Processing Systems}, volume~32. Curran Associates, Inc., 2019.
\newblock URL \url{https://proceedings.neurips.cc/paper_files/paper/2019/file/5103c3584b063c431bd1268e9b5e76fb-Paper.pdf}.

\bibitem[Salmanpour et~al.(2023)Salmanpour, Hosseinzadeh, Bakhtiyari, Maghsudi, and Rahmim]{Salmanpour2023}
Mohammad~R. Salmanpour, Mahdi Hosseinzadeh, Mahya Bakhtiyari, Mehdi Maghsudi, and Arman Rahmim.
\newblock Prediction of drug amount in parkinson's disease using hybrid machine learning systems and radiomics features.
\newblock \emph{International Journal of Imaging Systems and Technology}, 33:\penalty0 1437--1449, 7 2023.
\newblock ISSN 10981098.
\newblock \doi{10.1002/ima.22868}.

\bibitem[Shafer and Vovk(2008)]{Shafer2008}
Glenn Shafer and Vladimir Vovk.
\newblock A tutorial on conformal prediction.
\newblock Technical report, 2008.

\bibitem[Shamir et~al.(2015)Shamir, Dolber, Noecker, Walter, and McIntyre]{Shamir2015}
Reuben~R. Shamir, Trygve Dolber, Angela~M. Noecker, Benjamin~L. Walter, and Cameron~C. McIntyre.
\newblock Machine learning approach to optimizing combined stimulation and medication therapies for parkinson's disease.
\newblock \emph{Brain Stimulation}, 8:\penalty0 1025--1032, 11 2015.
\newblock ISSN 18764754.
\newblock \doi{10.1016/j.brs.2015.06.003}.

\bibitem[Sharma et~al.(2010)Sharma, Bachmann, and Linazasoro]{sharma2010classifying}
JC~Sharma, Cornelius~G Bachmann, and Gurutz Linazasoro.
\newblock Classifying risk factors for dyskinesia in parkinson’s disease.
\newblock \emph{Parkinsonism \& related disorders}, 16\penalty0 (8):\penalty0 490--497, 2010.

\bibitem[Stocchi et~al.(2024)Stocchi, Bravi, Emmi, and Antonini]{Stocchi2024}
Fabrizio Stocchi, Daniele Bravi, Aron Emmi, and Angelo Antonini.
\newblock Parkinson disease therapy: current strategies and future research priorities, 2024.
\newblock ISSN 17594766.

\bibitem[Sundararajan and Najmi(2020)]{Sundararajan2020}
Mukund Sundararajan and Amir Najmi.
\newblock The many shapley values for model explanation.
\newblock Technical report, 2020.

\bibitem[Svensson et~al.(2018)Svensson, Aniceto, Norinder, Cortes-Ciriano, Spjuth, Carlsson, and Bender]{Svensson2018}
Fredrik Svensson, Natalia Aniceto, Ulf Norinder, Isidro Cortes-Ciriano, Ola Spjuth, Lars Carlsson, and Andreas Bender.
\newblock Conformal regression for quantitative structure-activity relationship modeling - quantifying prediction uncertainty.
\newblock \emph{Journal of Chemical Information and Modeling}, 58:\penalty0 1132--1140, 5 2018.
\newblock ISSN 1549960X.
\newblock \doi{10.1021/acs.jcim.8b00054}.

\bibitem[Tibshirani et~al.(2019)Tibshirani, Barber, Candes, and Ramdas]{Tibshirani2019}
Ryan~J Tibshirani, Rina~Foygel Barber, Emmanuel Candes, and Aaditya Ramdas.
\newblock Conformal prediction under covariate shift.
\newblock \emph{Advances in neural information processing systems}, 32, 2019.

\bibitem[Tomlinson et~al.(2010)Tomlinson, Stowe, Patel, Rick, Gray, and Clarke]{Tomlinson2010}
Claire~L. Tomlinson, Rebecca Stowe, Smitaa Patel, Caroline Rick, Richard Gray, and Carl~E. Clarke.
\newblock Systematic review of levodopa dose equivalency reporting in parkinson's disease.
\newblock \emph{Movement Disorders}, 25:\penalty0 2649--2653, 11 2010.
\newblock ISSN 08853185.
\newblock \doi{10.1002/mds.23429}.

\bibitem[Vovk(2015)]{Vovk2015}
Vladimir Vovk.
\newblock Cross-conformal predictors.
\newblock \emph{Annals of Mathematics and Artificial Intelligence}, 74:\penalty0 9--28, 6 2015.
\newblock ISSN 10122443.
\newblock \doi{10.1007/s10472-013-9368-4}.

\bibitem[Vovk et~al.(2005)Vovk, Gammerman, and Shafer]{Vovk2005}
Vladimir Vovk, Alexander Gammerman, and Glenn Shafer.
\newblock \emph{Algorithmic Learning in a Random World}.
\newblock Springer, 2005.

\bibitem[Vovk et~al.(2018)Vovk, Nouretdinov, Manokhin, and Gammerman]{Vovk2018}
Vladimir Vovk, Ilia Nouretdinov, Valery Manokhin, and Alexander Gammerman.
\newblock Cross-conformal predictive distributions.
\newblock In \emph{conformal and probabilistic prediction and applications}, pages 37--51, 2018.

\bibitem[Watts et~al.(2021)Watts, Khojandi, Vasudevan, Nahab, and Ramdhani]{Watts2021}
Jeremy Watts, Anahita Khojandi, Rama Vasudevan, Fatta~B. Nahab, and Ritesh~A. Ramdhani.
\newblock Improving medication regimen recommendation for parkinson’s disease using sensor technology.
\newblock \emph{Sensors}, 21, 5 2021.
\newblock ISSN 14248220.
\newblock \doi{10.3390/s21103553}.

\bibitem[Wu et~al.(2019)Wu, Chen, Zhang, Xiong, Lei, and Deng]{Wu2019}
Jia Wu, Xiu~Yun Chen, Hao Zhang, Li~Dong Xiong, Hang Lei, and Si~Hao Deng.
\newblock Hyperparameter optimization for machine learning models based on bayesian optimization.
\newblock \emph{Journal of Electronic Science and Technology}, 17:\penalty0 26--40, 3 2019.
\newblock ISSN 2666223X.
\newblock \doi{10.11989/JEST.1674-862X.80904120}.

\bibitem[Ying(2019)]{Ying2019}
Xue Ying.
\newblock An overview of overfitting and its solutions.
\newblock In \emph{Journal of Physics: Conference Series}, volume 1168. Institute of Physics Publishing, 3 2019.
\newblock \doi{10.1088/1742-6596/1168/2/022022}.

\bibitem[Yuan et~al.(2021)Yuan, Beaulieu-Jones, Krolewski, Palmer, Veyrat-Follet, Frau, Cohen, Bozzi, Cogswell, Kumar, Coulouvrat, Leroy, Fischer, Sardi, Chandross, Rubin, Wills, Kohane, and Lipnick]{Yuan2021}
William Yuan, Brett Beaulieu-Jones, Richard Krolewski, Nathan Palmer, Christine Veyrat-Follet, Francesca Frau, Caroline Cohen, Sylvie Bozzi, Meaghan Cogswell, Dinesh Kumar, Catherine Coulouvrat, Bruno Leroy, Tanya~Z. Fischer, S.~Pablo Sardi, Karen~J. Chandross, Lee~L. Rubin, Anne~Marie Wills, Isaac Kohane, and Scott~L. Lipnick.
\newblock Accelerating diagnosis of parkinson’s disease through risk prediction.
\newblock \emph{BMC Neurology}, 21, 12 2021.
\newblock ISSN 14712377.
\newblock \doi{10.1186/s12883-021-02226-4}.

\end{thebibliography}

\newpage
\appendix
\section {Extended Background in Parkinson's Disease}
\label{sec:extended_pd}

PD is characterized by progressive degeneration of dopaminergic neurons with remarkable heterogeneity in clinical presentation, progression rates, and treatment responses. This heterogeneity presents significant challenges for medication management, as noted in studies by \citep{Jankovic1990} and \citep{Fereshtehnejad2015} who identified distinct PD subtypes with varying progression trajectories and treatment needs. Neurologists must navigate this complexity when making treatment decisions, particularly while adjusting dopaminergic medications \citep{Fox2018}.

Levodopa remains the gold standard medication for PD symptom control. With evidence for sustained therapeutic benefit for PD first shown in 1967 by \citep{Cotzias1967}, it remains the most effective medication to manage motor symptoms. In the brain, levodopa is converted to dopamine via dopa decarboxylase, replacing the neurotransmitter that is deficient in PD. Clinicians often prescribe various complementary medications (dopamine agonists, MAO-B inhibitors, COMT inhibitors) to address specific symptom profiles, and contraindications, or to prevent the breaking down of dopamine. However, long-term levodopa therapy presents diverse challenges: with chronic exposure, patients can experience LID as a side effect of non-physiological, pulsatile dopamine receptor stimulation \citep{Olanow2006}. Unlike the steady dopamine levels in healthy brains, oral levodopa creates waves of dopamine that rise and fall throughout the day \citep{Jenner2008}.  

Dyskinesias are involuntary, almost dance-like movements that can be extremely debilitating for patients, negatively impacting their quality of life. Given the complexity of managing multiple antiparkinsonian medications with different mechanisms and potencies, clinicians might benefit from standardized methods to quantify and compare treatment regimens. The LEDD has emerged as a standardized metric for comparing antiparkinsonian medications across different drug combinations and formulations. Initially developed by \citep{Tomlinson2010} and refined by subsequent studies \citep{Joshi2010, Kukkle2024} the LEDD converts each medication to its equivalent levodopa dose using specific conversion factors (e.g., 100mg levodopa = 133mg entacapone = 1mg pramipexole). This standardization enables consistent tracking of overall dopaminergic therapy over time and facilitates comparison across diverse medication regimens. 

Current practice for medication adjustment relies heavily on periodic clinical assessments \citep{Postuma2015, Goetz2003, Goetz2008} (typically every 3-6 months), subjective patient reporting, and expertise. However, this approach can fail to capture the full complexity of disease progression and treatment response, resulting in dosing decisions based on limited information about long-term medication needs \citep{Dorsey2020}. Consequently, patients can experience insufficient symptom control and suboptimal medication management, particularly challenging given the narrow therapeutic window of levodopa. These challenges are magnified in inpatient settings, with studies reporting medication timing errors in over 70\% of PD inpatient admissions; such errors can lead to complications like extended hospital stays, increased risk of infections, higher healthcare costs, and motor complications like muscle rigidity, and falls \citep{Aminoff2011, Gerlach2011}. 

\section {Summary of Current State of the Literature}
\label{sec:lit_summary}
Table~\ref{tab:lit_table} shows the different approaches. While these studies represent important advances in applying machine learning to PD medication management, they collectively fail to address the critical need for longitudinal LEDD forecasting with quantified uncertainty. The limitations of existing approaches are particularly evident in their reliance on point estimates without robust uncertainty quantification. 

\begin{table}[t]
  \centering 
  \caption{Literature review summary comparing previous approaches in PD medication forecasting across three dimensions: longitudinal medication prediction, use of LEDD, and uncertainty quantification. Our work uniquely addresses all three areas.}
  \resizebox{\textwidth}{!}{%
    \begin{tabular}{llll}
    \toprule
      \textbf{Authors} & \textbf{Longitudinal Medication Prediction} & \textbf{Use of LEDD} & \textbf{Uncertainty Quantification} \\
      \midrule
      \citep{Belyaev2023} et al. & \textcolor{red}{X} & \textcolor{red}{X} & \textcolor{red}{X} \\ 
      \citep{Lusk2023} et al. & \textcolor{red}{X} & \textcolor{red}{X} & \textcolor{red}{X} \\ 
      \citep{Palacios2011} et al. & \textcolor{red}{X} & \textcolor{red}{X} & \textcolor{red}{X} \\ 
      \citep{Yuan2021} et al. & \textcolor{red}{X} & \textcolor{red}{X} & \textcolor{red}{X} \\ 
      \citep{Latourelle2017} et al. & \textcolor{red}{X} & \textcolor{red}{X} & \textcolor{red}{X} \\ 
      \citep{Riasi2024} et al. & \textcolor{green}{\checkmark} & \textcolor{red}{X} & \textcolor{red}{X} \\ 
      \citep{Gutowski2023} et al. & \textcolor{red}{X} (Simulated, short-term optimization) & \textcolor{red}{X} & \textcolor{red}{X} \\ 
      \citep{Salmanpour2023} et al. & \textcolor{green}{\checkmark} (Fragmented by Intervals) & \textcolor{green}{\checkmark} & \textcolor{red}{X} \\ 
      \citep{Shamir2015} et al. & \textcolor{red}{X} (Post-surgical focus only) & \textcolor{green}{\checkmark} & \textcolor{red}{X} \\ 
      \citep{Watts2021} et al. & \textcolor{red}{X} (Clustering/classification only) & \textcolor{green}{\checkmark} & \textcolor{red}{X} \\ 
      \citep{Kim2021} et al. & \textcolor{red}{X} (Short-term optimization) & \textcolor{green}{\checkmark} & \textcolor{red}{X} \\ 
      \citep{Jatain2021} et al. & \textcolor{red}{X} & \textcolor{red}{X} & \textcolor{red}{X} \\ 
      \bottomrule
    \end{tabular}
  }%
  \label{tab:lit_table} 
\end{table}

\section {Parkinson's Disease Medications}

Our cohort analysis included a comprehensive range of antiparkinsonian medications administered during inpatient stays, as detailed in Table~\ref{tab:pd_meds}. Levodopa formulations were available in various delivery mechanisms to optimize absorption and minimize motor fluctuations. Dopamine agonists directly stimulate dopamine receptors with varying receptor affinities and half-lives, providing alternative or complementary therapy to levodopa. Supplementary medication classes included MAO-B and COMT inhibitors, which extend levodopa's duration of action by inhibiting dopamine breakdown; anticholinergics, primarily used for tremor management; amantadine for dyskinesia control; and the adenosine A2A antagonist that provides symptomatic improvement without direct dopaminergic stimulation. This diverse pharmacological approach reflects the complex nature of PD symptom management, where medication selection and dosing require careful consideration of disease stage, symptom profile, and individual patient characteristics to balance efficacy against side effects.

\label{sec:pd_meds}

\begin{table}[t]
  \centering
  \caption{Classification of antiparkinsonian medications administered during the study, organized by pharmacological mechanism of action, showing the diverse medication classes used for PD symptom management.}
  \begin{tabular}{ll}
  \toprule
  \textbf{Drug Class} & \textbf{Medications} \\
  \midrule
  Levodopa Formulations & Carbidopa-Levodopa, Carbidopa-Levodopa ER, \\
                        & Carbidopa-Levodopa-Entacapone, \\
                        & Carbidopa-Levodopa Intestinal Gel (Duopa) \\
  Dopamine Agonists & Pramipexole, Pramipexole ER, Ropinirole, \\
                    & Rotigotine, Apomorphine, Bromocriptine, Cabergoline \\
  MAO-B Inhibitors & Rasagiline, Selegiline, Safinamide \\
  COMT Inhibitors & Entacapone, Tolcapone \\
  Anticholinergics & Benztropine, Trihexyphenidyl \\
  NMDA Receptor Antagonists & Amantadine, Amantadine ER \\
  Adenosine A2A Antagonist & Istradefylline \\
  \bottomrule
  \end{tabular}
  \label{tab:pd_meds}
\end{table}

\section {Mean Feature Importance for Classification Model}
\label{sec:feature_importance}

Future importance analysis revealed key clinical insights: Mean LEDD per visit emerged as the strongest predictor, while age was the second most important feature, aligning with known associations between age, PD progression and increased medication over time \citep{goetz1988risk}. Clinical interaction patterns (days since last visit, length of stay) and medication-specific features (carbidopa-levodopa and several dopamine agonists) were also significant predictors, validating our model's ability to capture clinically relevant factors.  

\begin{figure}[t]
   \centering 
   \includegraphics[width=3.0in]{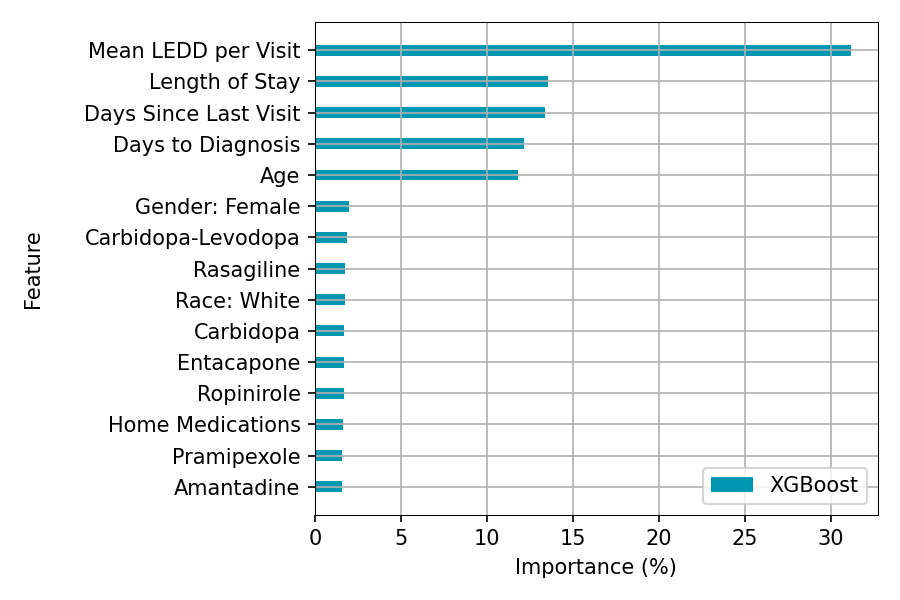}
   \caption{Mean feature importance for the classification model, showing that Mean LEDD per visit, Length of Stay, Days Since Last Visit, Days to diagnosis and Age as the strongest predictors for medication change.}
   \label{fig:fig4} 
 \end{figure}

\section{Conformal Prediction Framework}
\label{sec:cp_framework}
\paragraph{Naïve:} As described by \citep{Barber2021}, this approach evaluates conformity scores \textit{only on the training set}. For each observation, we calculated the conformity score: 
\begin{equation}\label{eq:eq9}
s_i = |y_i - \hat{y}_i|,
\end{equation}
and constructed prediction intervals as: 
\begin{equation}\label{eq:eq10}
C_\alpha(x) = \hat{y}(x) \pm \hat{q}_{1-\alpha},
\end{equation}
where $\hat{q}_{1-\alpha}$ is the $(1 - \alpha)$ empirical quantile of the training set conformity scores. 

\paragraph{Cross Validation Conformal (CV+):} This method employs K-fold cross-validation (K=5) to leverage multiple calibration sets. For each test point X, we aggregated predictions across folds to construct intervals as: 
\begin{equation}\label{eq:e11}
C_\alpha(x) = [\inf_j \hat{y}_j(x) - \hat{q}_{1-\alpha}^j, \sup_j \hat{y}_j(x) + \hat{q}_{1-\alpha}^j].
\end{equation}
\paragraph{Jackknife+ After Bootstrap (J+aB):}Designed specifically for ensemble methods that already use bootstrap samples. Instead of refitting models for each leave-one-out scenario, it reuses the bootstrap samples that don't contain a particular training point to create the leave-one-out predictions: 
\begin{equation}\label{eq:eq12}
C_\alpha(x) = [\hat{q}_{\alpha/2}^*(\{\hat{y}_{-i}^*(x)\}), \hat{q}_{1-\alpha/2}^*(\{\hat{y}_{-i}^*(x)\})].
\end{equation}

While both CV+ and J+aB provide the same theoretical coverage guarantee (at least $1 - 2\alpha$) without distributional assumptions, their implementation differences lead to distinct practical trade-offs. CV+ uses K distinct calibration sets with limited overlap, potentially offering more stable uncertainty estimates than the correlated leave-one-out predictions in J+aB, but at the cost of increased computation when not using ensemble methods. CV+ may occasionally exhibit over-coverage, producing intervals that are wider than necessary to achieve the target coverage level. 

J+aB offers remarkable computational efficiency for ensemble methods, as it reuses models already trained during the ensemble process, making it particularly suitable for large datasets or complex models where training is expensive. Its key advantage is providing valid prediction intervals ``for free'' with ensemble methods. However, the correlation between leave-one-out estimates may impact the stability of prediction intervals in smaller datasets. 

The Naïve method has no distribution-free coverage guarantee, exhibiting under-coverage in practice due to overfitting. However, it offers computational simplicity (requiring only one model fit) and often produces narrower intervals, which may be acceptable in applications where slight under-coverage is tolerable and computational resources are limited. 

\section {Calibration Analysis}
\label{sec:calib_analysis}

\begin{figure}[t]
   \centering 
   \includegraphics[width=3.5in]{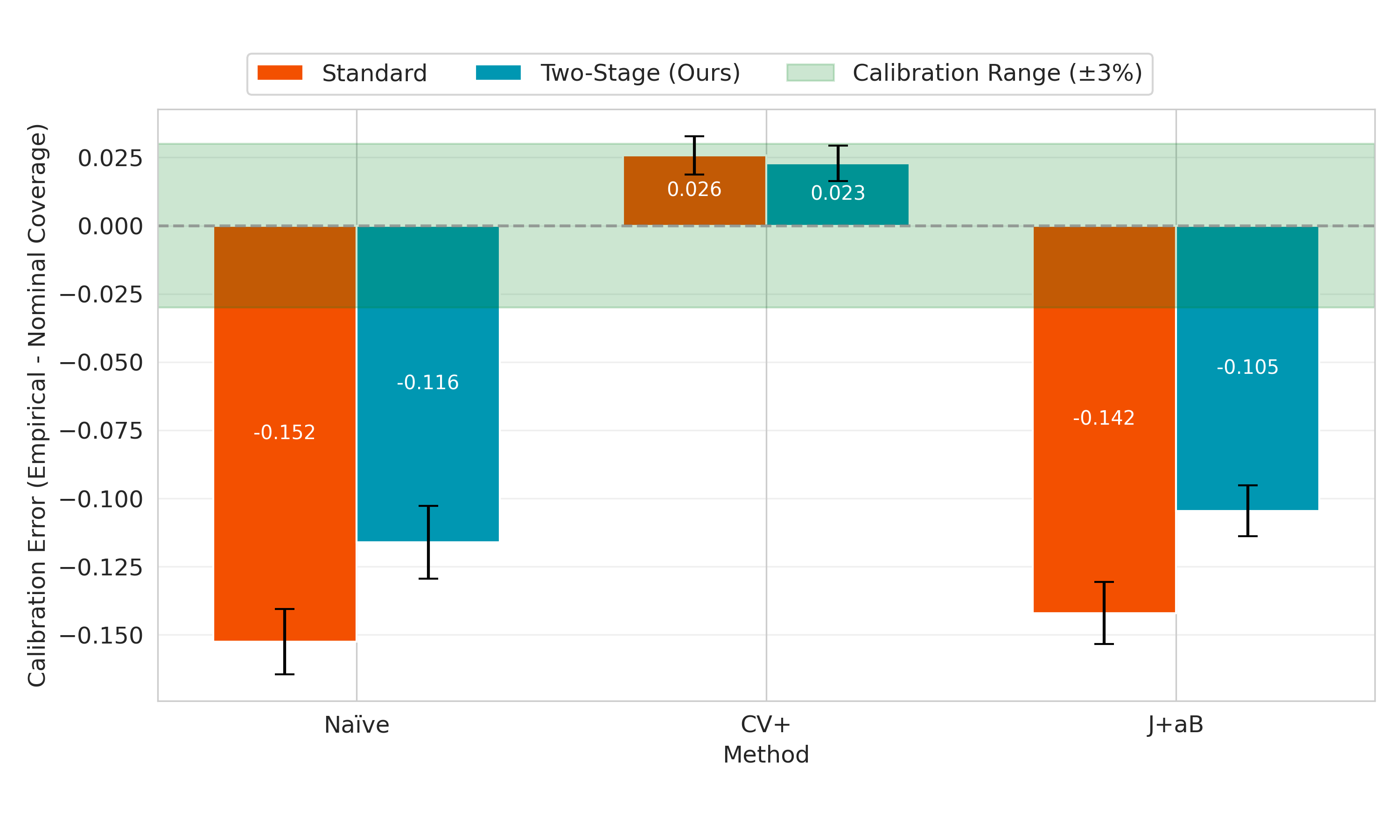} 
   \caption{Mean calibration error comparison between standard and two-stage approaches across three methods. The green area represents appropriate calibration range (±3\%), with CV+ showing proper calibration within this range.}
   \label{fig:mean_calib_error} 
 \end{figure}

We examined calibration quality by comparing nominal and empirical coverage. Figure~\ref{fig:mean_calib_error} quantifies these differences by showing the mean calibration error (Empirical - Nominal Coverage) for each method. The green area represents a practically acceptable calibration range (within ±3\% of nominal coverage), balancing statistical rigor with clinical utility. 

 Our two-stage CV+ approach demonstrates excellent calibration, with a mean calibration error of +2.3\%. In contrast, both Naïve and J+aB methods showed substantial under-coverage in both implementations, with Naïve exhibiting mean calibration errors of -15.2\% (standard) and -11.6\% (two-stage), while J+aB showed errors of -14.2\% (standard) and -10.5\% (two-stage). While our approach improves calibration across all methods compared to standard implementations (3.6\% calibration error improvement in Naïve, 3.7\% in J+aB), only CV+ achieves properly calibrated prediction intervals. 

  \begin{figure}[t]
   \centering 
   \includegraphics[width=4.5in]{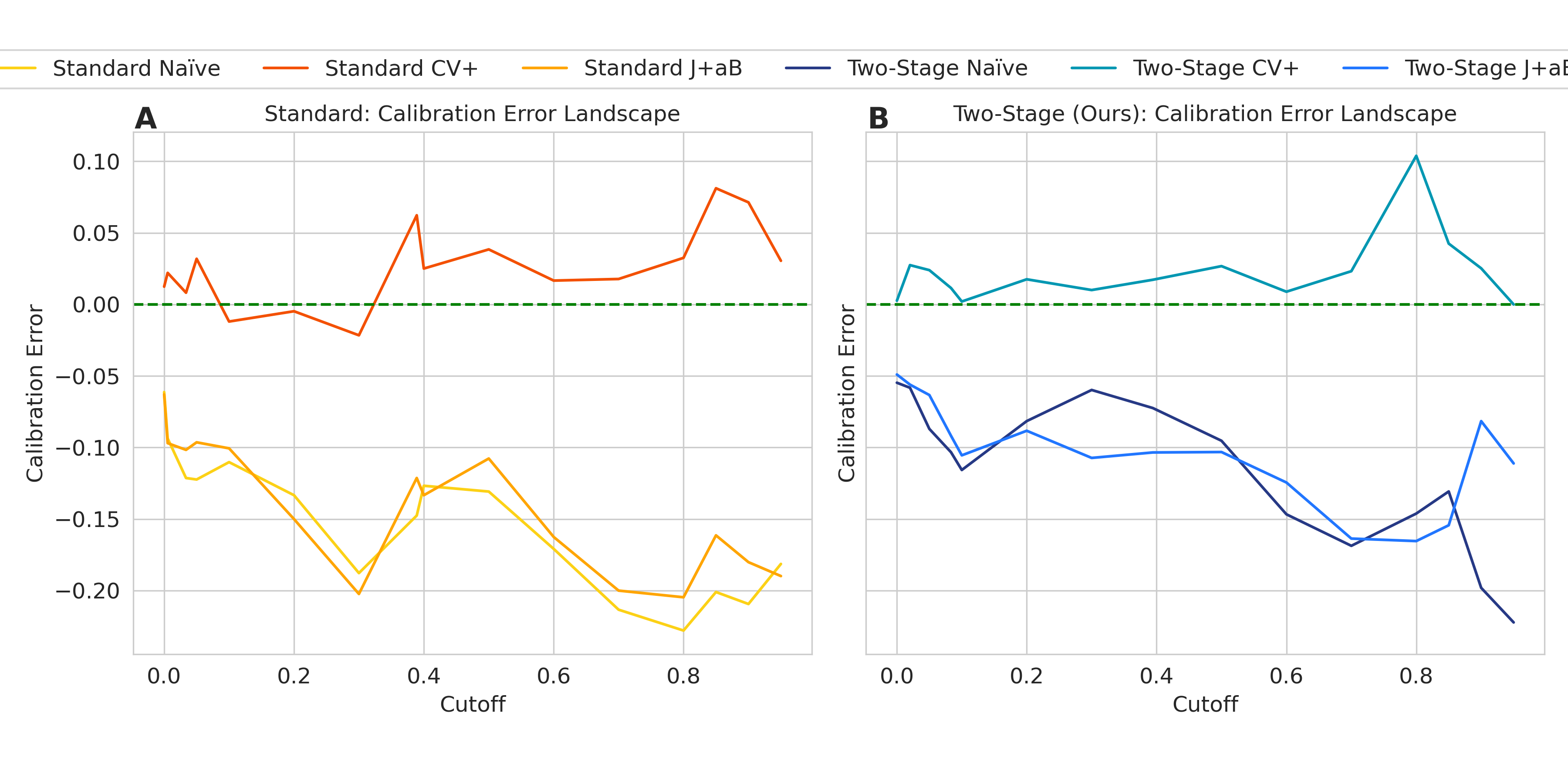} 
   \caption{Landscape calibration error across different classification cutoffs for standard (A) and two-stage (B) approaches. Values near zero indicate well-calibrated predictions, with CV+ maintaining better calibration throughout most cutoffs.}
   \label{fig:calib_error_landscape} 
 \end{figure}

Figure~\ref{fig:calib_error_landscape} provides deeper insight into this calibration behavior across different cutoffs, highlighting the reliability of prediction intervals, with values near zero (green dashed line) indicating well-calibrated predictions. For standard conformal prediction, in Figure~\ref{fig:calib_error_landscape}A, CV+ shows positive calibration errors between 0-8\%, maintaining conservative prediction intervals. In contrast, both Naïve and J+aB methods exhibit substantial under-coverage, with calibration errors becoming increasingly negative (reaching -20\% to -22\%) at higher cutoffs. This indicates that these methods may deliver significantly fewer correct predictions than their nominal guarantee suggests.  

In our two-stage approach (Figure~\ref{fig:calib_error_landscape}B), CV+ maintains calibration close to nominal coverage throughout most cutoffs, except for a brief spike around cutoff 0.8. Both Naïve and J+aB methods show substantially improved calibration compared to their standard conformal, though they still exhibit under-coverage (approximately -5\% to -15\%). 

These calibration results provide crucial context for our earlier findings on improved marginal coverage and mean interval length (Figure~\ref{fig:cp_comparison}C and Figure~\ref{fig:cp_comparison}D). The calibration of our two-stage CV+ approach demonstrates that its prediction intervals deliver correct statistical guarantees, making it suitable for clinical applications where dependable uncertainty quantification is essential. The improved but still imperfect calibration of Naïve and J+aB methods explains their occasionally narrower intervals. These methods achieve their relative precision by systematically producing intervals that contain the true LEDD value less frequently than their nominal rate suggests, particularly at longer time horizons.

\section {Conformal Prediction for Distribution Shift}
\label{sec:cp_distribution_shift}

\citep{Tibshirani2019} developed conformal prediction under covariate shift, adapting the framework to scenarios where training and test distributions differ, directly relevant to modeling progressive conditions like PD. Building on this foundation, \citep{Fannjiang2022} demonstrated practical applications of the methods for biomolecular design under feedback covariate shift, showcasing the method's adaptability to evolving systems. 

\citep{Prinster2023} introduced computationally efficient jackknife-based methods (JAW-FCS) that maintain rigorous coverage guarantees under feedback covariate shift while achieving practical balance between computational and statistical efficiency. In complementary work, \citep{Prinster2024} proved that conformal prediction can theoretically be extended beyond exchangeability assumptions to any joint data distribution, enabling valid uncertainty quantification even under AI/ML-agent-induced covariate shifts.

For heterogeneous PD populations, distributional conformal prediction \citep{Chernozhukov2021} offers even stronger guarantees by addressing heterogeneity in prediction uncertainty across different subpopulations. These methodological advances could substantially improve long-term prediction performance where our current methods demonstrate increasing uncertainty. 

\section {Statistical Analysis of Two-Stage Method Performance}
\label{sec:experiments}

We provide comprehensive statistical validation of our Two-Stage method's performance through 95\% bootstrap confidence intervals with 1000 resamples (Table~\ref{tab:bootstrap_descriptive}), and paired t-tests (Table~\ref{tab:statistical_tests}) to establish both statistical significance and practical importance of the observed improvements. We conducted 15 independent experimental runs comparing Standard and Two-Stage methods across three conformal prediction approaches (Naïve, CV+, J+aB) on our dataset.

\begin{table}[h]
\centering
\small 
\caption{Combined Performance Results with 95\% Bootstrap Confidence Intervals (1000 resamples), and Two-Stage performance changes for both coverage and interval length.}
\label{tab:bootstrap_descriptive}
\begin{tabular}{@{}lccccccc@{}}
\toprule
\multirow{2}{*}{Method} & \multicolumn{3}{c}{Coverage} & \multicolumn{3}{c}{Interval Length} \\
\cmidrule(lr){2-4} \cmidrule(lr){5-7}
 & Mean & 95\% CI & Two-Stage (\(\Delta\)) & Mean & 95\% CI & Two-Stage (\(\Delta\)) \\
\midrule
Naïve & 0.648 & [0.625, 0.671] & 0.684 (+5.6\%) & 0.316 & [0.243, 0.394] & 0.282 (+10.8\%) \\
CV+ & 0.826 & [0.812, 0.840] & 0.823 (-0.4\%) & 0.539 & [0.444, 0.632] & 0.421 (+21.8\%) \\
J+aB & 0.658 & [0.637, 0.680] & 0.695 (+5.7\%) & 0.413 & [0.344, 0.474] & 0.355 (+14.0\%) \\
\bottomrule
\end{tabular}
\end{table}

\begin{table}[h]
\centering
\caption{Paired T-Test Results and Effect sizes ($n=15$)}
\label{tab:statistical_tests}
\begin{tabular}{lcccccc}
\toprule
Comparison & t-statistic & p-value & Significance & Cohen's d\\
\midrule
Naïve Coverage & 3.150 & p = 0.007 & ** & 0.813 \\
CV+ Coverage & -0.258 & p = 0.800 & ns & -0.067 \\
J+aB Coverage & 4.048 & p = 0.001 & ** & 1.045 \\
Naïve Interval & 2.929 & p = 0.011 & * & 0.756 \\
CV+ Interval & 6.313 & p $<$ 0.001 & *** & 1.630 \\
J+aB Interval & 4.124 & p = 0.001 & ** & 1.065 \\
\bottomrule
\end{tabular}
\end{table}

From our analysis 5/6 comparisons show significant improvements ($p < 0.05$), with all methods achieving significant interval reductions while maintaining and/or improving coverage.

\section{Baseline Comparison}
\label{sec:baseline_comparison}

We compared our approach against a single-stage standard conformal prediction baseline. Cutoffs were selected based on optimal performance in Table~\ref{tab:cp_metrics_table}, specifically 0.95 for short to medium-term predictions (6M, 1Y, 2Y) and 0.85 for long-term predictions (4Y). Table~\ref{tab:baseline_comparison} demonstrates substantial improvements across all evaluated time horizons and methods. Standard conformal (without two-stage) shows severe under-coverage and inconsistent intervals across all timeframes.

\begin{table}[h]
\centering
\small
\caption{Baseline Comparison: Standard (Single-Stage) vs Two-Stage Conformal Prediction. Bolded pairs of coverage and interval length represent the best improvement per time window. Our approach provides reliable coverage (higher is better) with competitive interval lengths (lower is better).}
\label{tab:baseline_comparison}
\begin{tabular}{ll|cc|cc|cc}
\toprule
\textbf{Time} & \textbf{Method}
& \multicolumn{2}{c|}{\textbf{Standard (Single-Stage)}}
& \multicolumn{2}{c|}{\textbf{Two-Stage}}
& \multicolumn{2}{c}{\textbf{Improvement}} \\
\cmidrule(lr){3-4} \cmidrule(lr){5-6} \cmidrule(l){7-8}
& & \textbf{Coverage} & \textbf{Length}
& \textbf{Coverage} & \textbf{Length}
& \textbf{Coverage ($\Delta$)} & \textbf{Length ($\Delta$)} \\
\midrule
\multirow{3}{*}{6M}
& Naïve & 77.2\% & 0.219 & 67.0\% & 0.052 & -10.2 & -0.167 \\
& CV+ & 79.4\% & 0.264 & 87.3\% & 0.075 & \textbf{+8.0} & \textbf{-0.190} \\
& J+aB & 77.7\% & 0.239 & 75.1\% & 0.064 & -2.6 & -0.174 \\
\midrule
\multirow{3}{*}{1Y}
& Naïve & 73.3\% & 0.322 & 57.8\% & 0.041 & -15.5 & -0.281 \\
& CV+ & 81.1\% & 0.460 & 80.0\% & 0.082 & \textbf{-1.1} & \textbf{-0.378} \\
& J+aB & 75.0\% & 0.375 & 68.9\% & 0.122 & -6.1 & -0.254 \\
\midrule
\multirow{3}{*}{2Y}
& Naïve & 66.4\% & 0.226 & 59.6\% & 0.051 & -6.8 & -0.175 \\
& CV+ & 81.1\% & 0.811 & 90.4\% & 0.192 & \textbf{+9.2} & \textbf{-0.619} \\
& J+aB & 70.3\% & 0.703 & 71.2\% & 0.200 & +0.9 & -0.503 \\
\midrule
\multirow{3}{*}{4Y}
& Naïve & 53.6\% & 0.142 & 73.9\% & 0.063 & \textbf{+20.3} & \textbf{-0.080} \\
& CV+ & 84.0\% & 0.501 & 100.0\% & 0.685 & +16.0 & +0.184 \\
& J+aB & 62.4\% & 0.258 & 78.3\% & 0.355 & +15.9 & +0.097 \\
\bottomrule
\end{tabular}
\end{table}

\end{document}